\documentclass[lettersize,journal]{IEEEtran}
\usepackage{amsmath,amsfonts}
\usepackage{algorithmic}
\usepackage{algorithm}
\usepackage{array}
\usepackage[caption=false,font=normalsize,labelfont=sf,textfont=sf]{subfig}
\usepackage{textcomp}
\usepackage{stfloats}
\usepackage{url}
\usepackage{verbatim}
\usepackage{graphicx}
\usepackage{cite}
\usepackage[colorlinks=true, citecolor=green]{hyperref}
\usepackage{amssymb}
\usepackage{cleveref}
\usepackage{tabularray}
\usepackage{makecell}
\usepackage{float}
\usepackage{svg}
\usepackage{colortbl}
\usepackage{orcidlink}
\usepackage{multirow}
\usepackage{xcolor}

\begin{document}
\title{From Remote Sensing to Multiple Time Horizons Forecasts: Transformers Model for CyanoHAB Intensity in Lake Champlain}


\author{Muhammad~Adil\,\orcidlink{0009-0006-7435-0531},
        Patrick~J.~Clemins\,\orcidlink{0000-0002-7930-3025}, 
        Andrew~W.~Schroth\,\orcidlink{0000-0001-5553-3208}, 
        Panagiotis~D.~Oikonomou\,\orcidlink{0000-0001-6612-0994}, 
        Donna~M.~Rizzo\,\orcidlink{0000-0003-4123-5028}, 
        Peter~D.~F.~Isles\,\orcidlink{0000-0003-4446-6788}, 
        Xiaohan~Zhang\,\orcidlink{0000-0001-6344-9604},
        Kareem~I.~Hannoun\,\orcidlink{0000-0003-3176-1104}, 
        Scott~Turnbull\,\orcidlink{0000-0002-4384-652X},
        Noah~B.~Beckage\, \orcidlink{0009-0000-9026-9510},
        Asim~Zia\,\orcidlink{0000-0001-8372-6090}, 
        and~Safwan~Wshah\,\orcidlink{0000-0001-5051-7719}
      
\thanks{M. Adil, X. Zhang, and S. Wshah are with Vermont Artificial Intelligence Laboratory (VaiL),  Department of Computer Science, University of Vermont, Burlington, VT 05405 USA.}
\thanks{A. Zia is with Department of Community Development and Applied Economics \& 
Department of Computer Science, University of Vermont, Burlington, VT 05405 USA.}
\thanks{A. W. Schroth is with Department of Geology, University of Vermont, Burlington, VT 05401 USA.}
\thanks{P. D. Oikonomou and D. M. Rizzo are with Department of Civil and Environmental Engineering, University of Vermont, Burlington, VT 05405 USA.}
\thanks{P. J. Clemins, S. Turnbull, and N. B. Beckage are with Vermont EPSCoR (Established Program to Stimulate Competitive Research), University of Vermont, Burlington, VT 05405 USA.}
\thanks{P. D. F. Isles is with Vermont Department of Environmental Conservation, Montpelier, VT 05602 USA.}
\thanks{K. I. Hannoun is with Water Quality Solutions, Inc., McGaheysville, VA 22840 USA.}
\thanks{Corresponding Author: Safwan Wshah, e-mail: safwan.wshah@uvm.edu.}
\thanks{Manuscript received Month DD, YYYY; revised Month DD, YYYY.}}


\markboth{IEEE JOURNAL OF SELECTED TOPICS IN APPLIED EARTH OBSERVATIONS AND REMOTE SENSING, Vol.~XX, No.~X, Month~YEAR}%
{Author Adil \MakeLowercase{\textit{et al.}}: Transformers-based Deep Learning for CyanoHABs Prediction}


\maketitle

\begin{abstract}
Cyanobacterial Harmful Algal Blooms (CyanoHABs) pose significant threats to aquatic ecosystems and public health globally. Lake Champlain is particularly vulnerable to recurring CyanoHAB events, especially in its northern segment: Missisquoi Bay, St. Albans Bay, and Northeast Arm, due to nutrient enrichment and climatic variability.  Remote sensing provides a scalable solution for monitoring and forecasting these events, offering continuous coverage where in-situ observations are sparse or unavailable. In this study, we present a remote sensing-only forecasting framework that combines Transformers and BiLSTM to predict CyanoHAB intensities up to 14 days in advance. The system utilizes Cyanobacterial Index data from the Cyanobacterial Assessment Network and temperature data from Moderate Resolution Imaging Spectroradiometer satellites to capture long-range dependencies and sequential dynamics in satellite time series. The dataset is very sparse, missing more than 30\%  of the Cyanobacterial Index data and 90\% of the temperature data. A two-stage preprocessing pipeline addressed data gaps by applying forward fill and weighted temporal imputation at the pixel level, followed by smoothing to reduce the discontinuities of CyanoHAB events. The raw dataset is transformed into meaningful features through equal-frequency binning for the Cyanobacterial Index values and extracted temperature statistics. Transformer-BiLSTM model demonstrates strong forecasting performance across multiple horizons, achieving F1 scores of 89.5\%, 86.4\%, and 85.5\% at one-, two-, and three-day forecasts, respectively, and maintaining an F1 score of 78.9\% with an AUC of 82.6\% at the 14-day horizon. These results confirm the model’s ability to capture complex spatiotemporal dynamics from sparse satellite data and to provide reliable early warning for CyanoHABs management. The dataset and trained model weights are publicly available, facilitating further research in CyanoHABs forecasting. 
\end{abstract}


\section{Introduction} \label{sec:introduction}
Cyanobacterial harmful algal blooms (CyanoHABs) pose a significant environmental challenge that affects water quality, aquatic ecosystems, and public health in inland waters worldwide. CyanoHABs are characterized by the excessive growth of cyanobacteria in inland water systems, which produce toxins harmful to humans, ecological health, and the broader environment. Their formation is primarily attributed to eutrophication \cite{rolim2023remote}, and is exacerbated by anthropogenic activities such as intensive agriculture \cite{klemas2012remote, pamula2023remote} and insufficient policy actions to preserve the water quality of lake bodies \cite{roelke2011decade}. The combined effect of eutrophication and anthropogenic activities leads to an increase in nutrient fluxes, accompanied by lighting conditions, and water temperature favorable to cyanobacteria that can lead to more severe and extensive CyanoHABs \cite{roelke2011decade, thomas2003satellite, blondeau2014review}. CyanoHABs can drastically degrade inland-water quality \cite{antoniou2005Cyanotoxins, deng2017addressing}, threaten public health \cite{backer2002cyanobacterial, fleming2011review, brooks2016harmful}, decrease biodiversity, contribute to habitat loss \cite{landsberg2002effects}, which in turn causes damage to socioeconomic and ecological aspects of aquatic ecosystems \cite{sanseverino2016algal}. In freshwater systems, CyanoHABs species that produce biotoxins are of particular concern as these toxins can accumulate in aquatic food webs, when consumed by humans, leading to severe illness, paralysis, and even death \cite{berdalet2016marine}. According to a scientific research summary from  \cite{kudela2015harmful}, CyanoHABs cost \$4.6B per year in estimated damages within the United States alone. Early and reliable CyanoHABs forecasting systems can help mitigate adverse effects or even prevent them.

Numerous studies have identified temperature as a factor in the proliferation of CyanoHABs \cite{hecht2022modeling, gobler2017ocean, gobler2020climate, ho2020exploring, paerl2016duelling, cha2017relative, tang2022spatiotemporal, lapucci2022towards, igwaran2024cyanobacteria, marrone2023toward}. Other environmental factors include precipitation \cite{ho2020exploring, zhou2021effect, larsen2020extreme, haakonsson2017temperature}, nutrient concentration \cite{lemley2021harmful, wang2021harmful, glibert2020further}, wind conditions \cite{zhang2020view, silva2021twenty}, and hydraulic retention time \cite{keeler2012linking, carey2012eco, giani2020comparing}. Although these CyanoHAB drivers are well understood, early CyanoHAB forecasting remains challenging, as environmental factors vary spatially and temporally \cite{kim2022simultaneous}. 

Given these growing concerns, numerous CyanoHABs forecasting models have been developed in recent years. However, the majority of these algorithms rely on in-situ sensors \cite{kim2022simultaneous, lee2022integrated, villanueva2023one, liu2022algal} or manual water sampling \cite{silva2023forecasting, molares2023hybrid, torbick2015multiscale}, both of which are costly and labor-intensive. These data sources are also limited in their ability to capture large spatial areas, thereby constraining the detection of temporal and spatial changes in CyanoHABs, which are crucial for understanding their formation.

Remote sensing offers a promising solution to the limitations of sensor-based and manual sampling methods. Recent technological advances in remote sensing can transform the CyanoHABs monitoring landscape by providing data at higher temporal and spatial resolutions, enabling comprehensive and sophisticated modeling of the dynamic nature of CyanoHABs. 

Many researchers have used remote sensing to forecast and predict harmful algal blooms across aquatic systems over the past decade \cite{steidinger1981biologic,izadi2021remote, manning2019extending, gupta2023sub, matthews2023near}. Early demonstrations  \cite{steidinger1981biologic} focused on marine red tides using data from the Coastal Zone Color Scanner (CZCS) aboard Nimbus-7 in 1981. Although biologically distinct from freshwater cyanobacterial blooms, this pioneering work established the remote-sensing foundations later adapted for inland water monitoring and influenced the development of sensors such as  Moderate Resolution Imaging Spectroradiometer (MODIS), Medium Resolution Imaging Spectrometer (MERIS), and the Sentinel satellites. Subsequent work applied these sensors to freshwater CyanoHAB forecasting using a range of machine-learning and statistical methods. \cite{izadi2021remote} utilized the MODIS data with XGBoost, Random Forest, and Support Vector Machines to predict CyanoHAB occurrences with lead times of 1 to 11 days. A sub-monthly CyanoHAB forecasting model was developed by incorporating remote sensing and environmental factors to predict CyanoHAB cell counts \cite{gupta2023sub}. Satellite-derived chlorophyll‑\emph{a} data were combined with univariate time-series models, enabling 1 to 4 weeks of CyanoHABs forecast across 15 lakes worldwide \cite{matthews2023near}. A regression-based model was developed to predict the occurrence, spatial distribution, and intensity of CyanoHABs using chlorophyll‑\emph{a} concentration imagery from MODIS satellites \cite{manning2019extending}. 

Although these studies offer significant improvements for CyanoHABs detection and forecasting, they have at least one of the following drawbacks: 1) Limited transferability due to a reliance on ground sensor data in addition to remote sensing data, restricting the model's scalability to regions without ground sensing data; and 2) Most of these studies do not include CyanoHABs intensities. Predicting the intensity of CyanoHABs has a significant role in effective environmental management and public health protection, and early awareness helps inform public policy and protect human health.

To address the aforementioned limitations, we present a Transformer-BiLSTM deep learning framework that combines a multi-head, self-attention Transformer \cite{vaswani2017attention} and Bidirectional Long Short-Term Memory (BiLSTM) architectures for extended-horizon CyanoHAB intensity forecasting, utilizing exclusively remote sensing data. The proposed Transformer-BiLSTM model combines the global context modeling capability of Transformers with BiLSTM, further refining Transformers’ contextual understanding by maintaining temporal continuity and smoothing short-term variations. Unlike prior approaches that are constrained to in situ data, the proposed framework relies solely on satellite-derived inputs. To overcome the inherent data sparsity challenges in satellite observations, the framework employs a two-stage preprocessing pipeline that combines temporal imputation and smoothing techniques. The model generates forecasts across five distinct intensity classes (Low, Medium, High, Very High, and Extreme Intensity CyanoHABs) with forecast horizons of 14 days. This approach effectively predicts the likelihood and severity of CyanoHAB events across multiple time scales. Fine-grained intensity forecasting with an extended lead time provides environmental agencies and public health authorities with timely and actionable insights for proactive risk assessment and response planning.

We summarize our contributions in six principal ways:
\begin{itemize}

    \item We introduce a hybrid Transformer-BiLSTM architecture that integrates a multi-head, self-attention transformer and BiLSTM architectures for predicting the occurrence and intensity of CyanoHABs, enabling accurate forecasting up to 14 days, bridging short and medium‑range early‑warning needs.
    
    \item We formulate the task as a bin‑wise, five‑level intensity prediction, allowing managers to assess not only whether a bloom will occur but also its expected severity.  

    \item Our approach exclusively utilizes remote sensing data for CyanoHAB forecasting, overcoming sensor-based limitations. To our knowledge, this is the first model to rely entirely on remote sensing for multi-horizon CyanoHAB incidence and intensity prediction.
    
    \item This framework represents the first CyanoHAB early-warning system for Lake Champlain and is transferable to other Contiguous United States freshwater systems using freely available remote sensing data with minimal re-training.
    
    \item We make our processed remote sensing dataset and preprocessing pipeline publicly available to facilitate future CyanoHABs research.    

    \item The effectiveness of the Transformer-BiLSTM model is evaluated against Advanced Deep Learning and Machine Learning models. The proposed model demonstrates the superiority of CyanoHABs’ early forecasting and intensity prediction. 
    
\end{itemize}

\section{Related Work}  \label{sec:related_work}
\subsection{In-Situ Based CyanoHABs Forecasting}
In-situ based CyanoHABs prediction relies on in-situ field monitoring and is sometimes coupled with automated sensors. These approaches often use parameters such as chlorophyll‑\emph{a} concentrations, dissolved oxygen, nutrient fluxes, water temperature, pH, and cyanobacteria cell counts to predict CyanoHAB occurrence or intensity. The research investigating the detrimental effects of excessive nutrients, specifically nitrogen and phosphorus, related increased CyanoHAB activity to the exceedance of a critical N:P molar ratio \cite{wang2021harmful}. A 7-day forecasting model utilizing microcystin concentration as the primary indicator and incorporating input features such as pH, precipitation, temperature, dew point, and wind speed was developed to predict CyanoHABs in Iowa lakes \cite{villanueva2023one}. A hybrid deep learning approach combining wavelet analysis with LSTM was developed to forecast the CyanoHABs in Lake Mendota \cite{liu2022algal}. The research used hourly, daily, and monthly chlorophyll concentrations and cyanobacterial cell biomass counts to forecast CyanoHABs at hourly, daily, and monthly resolutions. Boosted regression trees and artificial neural networks are used to investigate the impact of high-temporal-resolution physicochemical and meteorological data features on the growth and decline of CyanoHABs in Upper Klamath Lake, Oregon. The authors concluded that decreases in inflows and lake-surface elevation, and increases in temperature and phosphorus concentration, play a significant role in CyanoHAB growth. Machine learning models for forecasting CyanoHABs in coastal areas were studied using physical, geochemical, and climate data to predict chlorophyll concentration as a proxy for CyanoHABs in Biscayne Bay \cite{yan2024predicting}.

Recent works have begun incorporating attention mechanisms into CyanoHABs forecasting \cite{zhang2025enhancing, kim2023incorporation, ahn2023ensemble}. Reverse‑time and dual‑stage attention mechanisms are combined with recurrent imputation to achieve 1, 7, and 14-day CyanoHABs forecasts at Nakdong River in South Korea \cite{kim2023incorporation}.  Stacked gradient‑boosting trees with an attention‑based CNN‑LSTM demonstrated that attention consistently boosted ensemble skill over purely boosting or purely deep learning baselines for CyanoHABs forecasting \cite{ahn2023ensemble}. Most recently, a Temporal Convolution Network fused with multi‑head attention and BiLSTM was developed to predict hour‑ahead chlorophyll‑\emph{a} from high‑frequency buoy measurements \cite{zhang2025enhancing}. While these works confirm the value of attention for capturing abrupt fluctuations, they remain sensor‑bound and focus on either binary CyanoHABs status or chlorophyll proxies at a single location. 

Given our focus on the Lake Champlain study area, reviewing previous CyanoHABs monitoring and forecasting studies conducted in this region is essential. A modified self-organizing map was used to analyze relationships between cyanobacteria blooms and environmental conditions in Lake Champlain's Missisquoi Bay \cite{pearce2013unraveling}. The authors concluded that low dissolved nitrogen-to-soluble reactive phosphorus ratios and sediment anoxia strongly correlate with bloom conditions. Bowling et al. A 9-year phytoplankton dataset from Missisquoi Bay was analyzed using water samples collected from four sites through manual field sampling methods, laboratory analysis, and meteorological data, showing weak correlations between environmental factors and cyanobacterial distributions, with the heterogeneity of bloom patterns remaining largely unexplained \cite{bowling2015heterogeneous}. High-frequency sensor data, field sampling, and long-term monitoring data were utilized to investigate the internal drivers of cyanobacteria blooms in Missisquoi Bay \cite{isles2015dynamic}. The authors found that internal phosphorus loading, thermal stratification, and shifting nutrient limitations play a vital role in bloom progression.

\subsection{Remote Sensing CyanoHABs Detection}
Remote sensing can capture the temporal and spatial variability of CyanoHABs. Extensive literature \cite{izadi2021remote, pamula2023remote, manning2019extending, hill2019habnet, wynne2024remote, liu2022remote, myer2020spatio, gupta2023sub, matthews2023near} describes the monitoring, detection, and forecasting of CyanoHABs using remote sensing. MERIS, Sentinel-2, Sentinel-3, and satellites equipped with moderate-resolution imaging spectroradiometers (MODIS-Aqua and MODIS-Terra) are transforming the CyanoHABs monitoring and modeling landscape by providing higher-frequency, open-source data. They can capture detailed images and information about water bodies, enabling researchers to effectively detect and track the development of CyanoHABs.  

Four machine learning models were compared for forecasting CyanoHABs in Lake Erie at 10-, 20-, and 30-day lead times \cite{gupta2023sub}. The author used satellite-derived CI data alongside meteorological, hydrodynamic, and nutrient data as inputs to predict future CI values as a proxy for CyanoHABs intensity. 
A. Gupta \cite{gupta2023sub} compared four machine learning models for forecasting harmful algal blooms in Lake Erie at 10-, 20-, and 30-day lead times.  A simple, generalizable univariate forecasting model was developed to predict cyanobacterial blooms using satellite-derived chlorophyll‑\emph{a} data \cite{matthews2023near}. Forecasts were generated at 1-, 2-, and 4-week horizons for 15 lakes worldwide. The 1-week model accurately predicted high-risk blooms with 80\% accuracy, comparable to that of complex models. Landsat-8 and Sentinel-2 satellite imagery were used to estimate chlorophyll‑\emph{a}, turbidity, and phycocyanin concentrations using support vector and random forest regression models \cite{pamula2023remote}. The impact of spatial resolution on the detection of CyanoHABs was investigated using data from Landsat-8, Sentinel-2, and PlanetScope imagery \cite{liu2022remote}. The authors evaluated twenty bio-optical algorithms from earlier studies to predict chlorophyll concentration, biomass, phyco-cyanin concentration, etc. A spatiotemporal modeling approach utilizing Bayesian hierarchical models was presented to predict the likelihood of CyanoHABs in Florida's freshwater systems \cite{myer2020spatio}. The authors used weekly composite images of maximum cyanobacteria abundance from Sentinel-3 imagery, along with ambient temperature, surface water temperature, and precipitation data. A collaborative framework for monitoring and predicting harmful algal bloom accumulation in nearshore areas of Lake Chaohu was developed using a coupled hydrodynamic-water quality-algae model \cite{qiu2023development}. The authors integrated satellite remote sensing data, 42 land-based video monitoring devices providing hourly coverage, and in-situ buoy stations measuring chlorophyll‑\emph{a} and lake currents. The framework provided real-time quantitative monitoring and daily predictions of harmful algal bloom accumulation risks in nearshore zones to support emergency lake management.  An extreme gradient boosting algorithm was used to estimate chlorophyll concentrations as an indicator of CyanoHABs by training on in-situ chlorophyll measurements and using reflectance indices from Landsat-8 Operational Land Imager (OLI) data along with other environmental inputs \cite{cao2020machine}. Recently, \cite{qiu2025monitoring} presented a multi-source monitoring and prediction framework for CyanoHABs, integrating satellite, UAV, ground-based, and in-situ systems to enable high-frequency, nearshore-focused assessment in Lake Chaohu. By simulating the growth-drift-accumulation dynamics, the presented method significantly improves prediction accuracy and practical early warning capabilities in eutrophic lakes.

Lake Champlain has been the subject of numerous CyanoHAB studies. Empirical regression models were used to map in-situ data with corresponding Landsat bands and blue, green, and red band ratios \cite{trescott2012remote}. Pigment concentration retrieval methods were evaluated in Missisquoi Bay, Lake Champlain, demonstrating that empirically calibrated QuickBird data using NIR/Red band ratios could explain approximately 80\% of chlorophyll‑\emph{a} variability while achieving an $R^2$ of 0.68 for Phycocyanin concentration mapping \cite{wheeler2012mapping}. CyanoHABs in Lake Champlain were mapped using Landsat-8 Operational Land Imager, Rapid Eye, and Proba Compact High-Resolution Imaging Spectrometer \cite{torbick2015multiscale}. CyanoHABs were detected using band ratios to identify spatial patterns within the lake and problematic bays with high phycocyanin concentrations. MERIS and CI are used to estimate the cyanobacteria cell counts (cells/mL) across eastern United States (US) lakes \cite{lunetta2015evaluation}. The authors identified good performance at low (10,000 to 109,000) and high ($>$ 1,000,000) cell counts; however, the approach performed suboptimally at intermediate concentrations. An ensemble-based system using down-scaled MODIS imagery was developed to estimate chlorophyll‑\emph{a} concentrations in inland lakes \cite{8736492}. The system combines adaptive modeling and Gaussian quadrature to minimize classification and estimation errors, thereby enhancing accuracy across bloom stages. Recently, the variability of CyanoHABs in Missisquoi Bay, Lake Champlain, was analyzed using satellite imagery from OLCI and MODIS with the CI algorithm to assess temporal and spatial occurrences of CyanoHABs \cite{wynne2024remote}.

While previous studies have enhanced our understanding of CyanoHABs' spatial and temporal distribution in Lake Champlain, forecasting is limited due to reliance on in situ sensor data and the limited temporal resolution of site-specific remote sensing data. This dependence on location-specific measurements restricts the transferability of the model to other water bodies. Developing advanced forecasting systems based exclusively on widely available remote sensing data is crucial for creating models that can be deployed across diverse freshwater ecosystems with minimal adaptation.

\section{Study Area} \label{sec:study_area}

\begin{figure*}[!t]
\centering
\includegraphics[width=1\textwidth, height=1\textwidth, keepaspectratio]{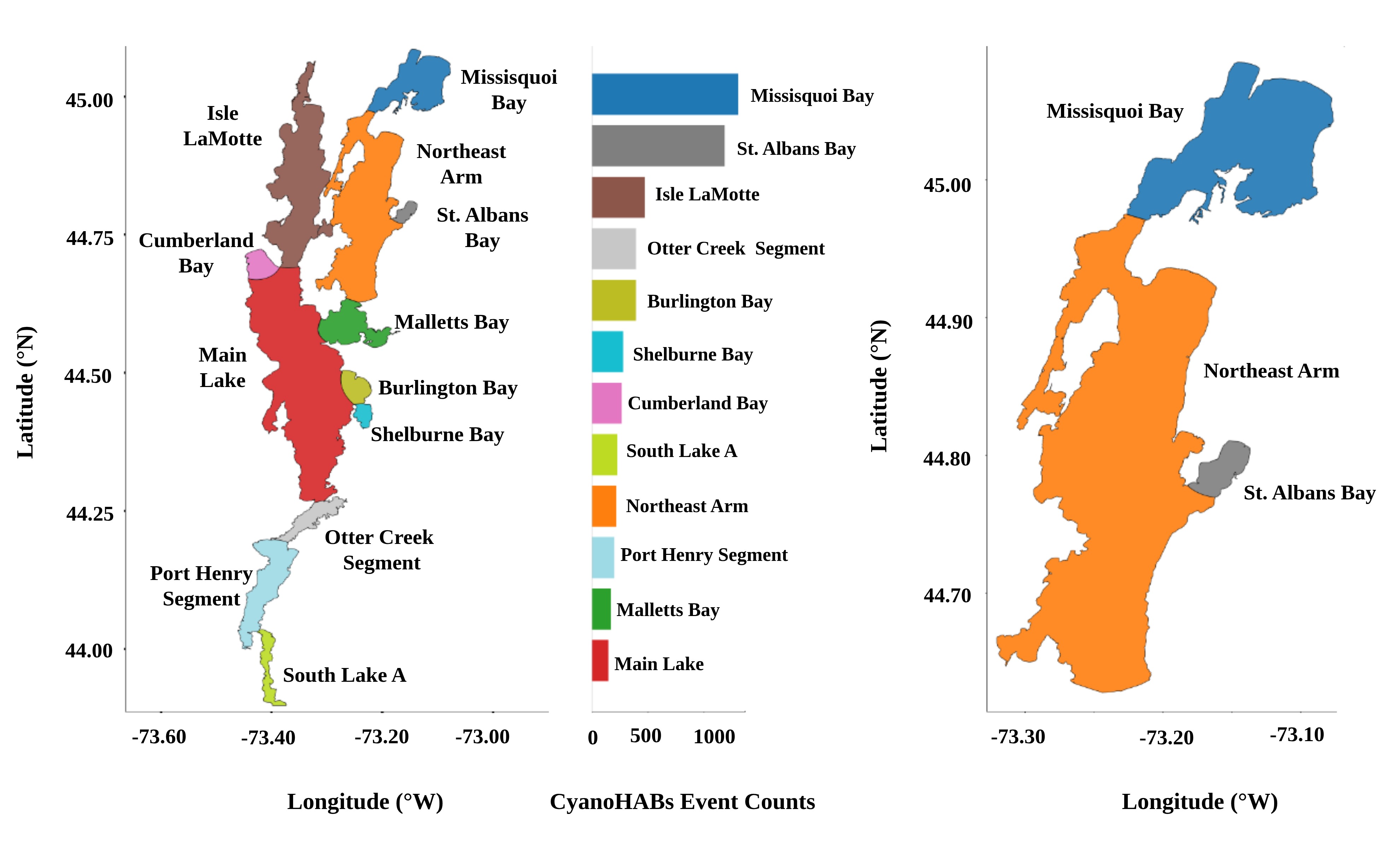}
\caption{The visual illustration of Lake Champlain shows 12 segments and their CyanoHABs events as a Bar Chart on the right side from 2016 to 2023. We can see that Missisquoi Bay and St. Albans Bay experienced an extensive number of CyanoHABs during that duration. The right side map shows the three stations - Missisquoi Bay, St. Albans Bay, and Northeast Arm - used in the study.}
\label{lake_champlain_overview}
\end{figure*}

Lake Champlain is a glacially-formed lake that consists of more than 70 islands and 54 beaches with annual economic revenues of \$580 million \cite{vaughan2021lake, decerega2016value, LCBPFacts2023}. Lake Champlain is the 13th-largest lake in the US, situated between New York and Vermont, and extending into Canada to the North. The lake is 120 miles long with a surface area of 1127 km$^2$ and a volume of 25.8 km$^3$ \cite{epa_lake_champlain}. It is 19km at its widest point, with an estimated maximum and mean depth of 122m and 23m, respectively \cite{LCBPFacts2023}. With more than 800km of shoreline and $21000$ km$^{2}$ in watershed area, it supports an approximate population of 571,000 \cite{vaughan2021lake}. Despite its relatively small size, the lake drainage basin is quite significant in comparison, with a drainage ratio of 19:1 \cite{facey2012lake}. The large drainage basin makes Lake Champlain susceptible to severe precipitation events and snowmelt, with runoff from agriculture, industry, and urbanization all contributing to eutrophication, which subsequently leads to the growth of CyanoHABs. \Cref{lake_champlain_overview} shows the approximate number of CyanoHABs events, at each of the 12 Lake Champlain monitoring segments from January 2016 to November 2023.

This study focuses on three lake segments: Missisquoi Bay, St. Albans Bay, and the Northeast Arm. The Northeast Arm is categorized as mesotrophic, indicative of relatively low nutrient levels and biological productivity. It makes up about one-quarter of Lake Champlain's surface area and has a mean depth of 13m. Missisquoi Bay and St. Albans Bay are eutrophic zones with elevated nutrient concentrations and biological productivity. These areas, consequently, experience the highest annual incidences of CyanoHABs, as reported in \Cref{lake_champlain_overview}. Both bays - Missisquoi Bay and St. Albans Bay - are shallow and warm, providing favorable conditions for CyanoHABs growth. Missisquoi Bay has a mean depth of 2.5m, while St. Albans Bay has a mean depth of 8m \cite{levine2012eutrophication}.  

\begin{figure*}[h]
\centering
\includegraphics[width=0.95\textwidth, height=1\textwidth, keepaspectratio]{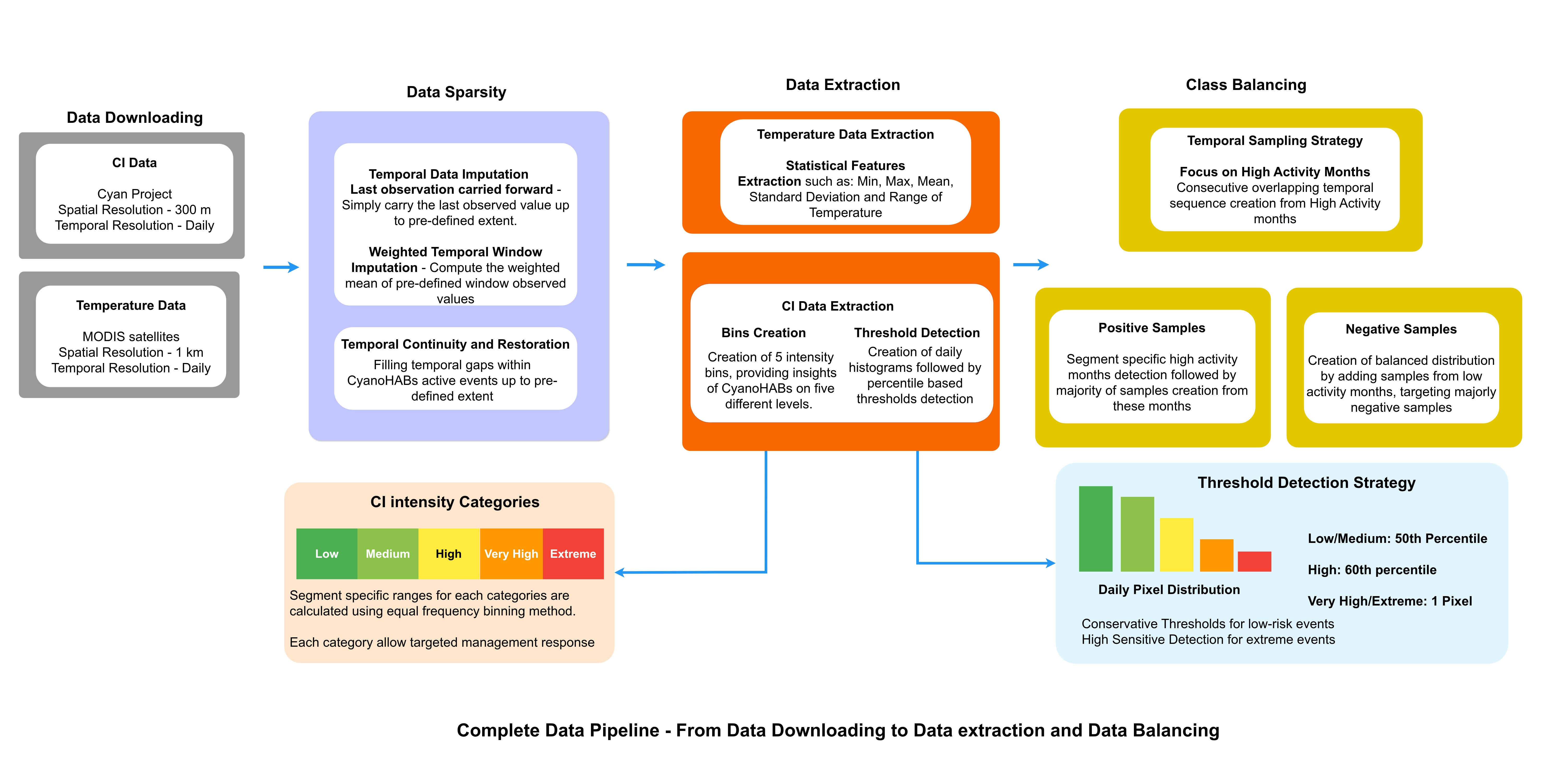}
\caption{Complete data processing pipeline for CyanoHABs prediction, from data acquisition through class balancing. The workflow consists of four stages: data downloading (CI and temperature), data sparsity management through imputation, data extraction for feature engineering, and class balancing using temporal sampling strategies.}
\label{data_pipeline}
\end{figure*}

\section{Data Collection \& Processing}
We use satellite remote sensing time series from two sources: the Cyanobacterial Index from the Cyanobacterial Assessment Network (CyAN) \cite{OBPG_CyAN_CI_v6_2024}, derived from ocean-color satellite imagery at 300 m spatial resolution (daily and 7-day maximum composites), and Moderate Resolution Imaging Spectroradiometer (MODIS) \cite{hulley2018viirs} land surface temperature (LST) at 1 km with separate day/night daily retrievals. The study period spans January 2016 to November 2024 for three Lake Champlain segments - Missisquoi Bay, St. Albans Bay, and the Northeast Arm. All inputs are in the form of GeoTiff, where shallow water pixels are masked, and aggregated to segment–day records with statistical features extracted from LST imagery. Further details are provided in the next four subsections:1) Data Collection - downloading CI data from CyAN and temperature data from MODIS, 2) Data sparsity handling - addressing substantial data sparsity through temporal imputation methods, 3) Data Extraction - extracting model ready features with five CI intensity categories and ten temperature statistics, 4) Class Balancing - temporal sampling for class balancing to focus on high activity months with later addition of negative samples from inactive months. \Cref{data_pipeline} illustrates the complete data pipeline steps visually.

\subsection{Data Collection}
\subsubsection{Cyanobacterial Index Values}
The Cyanobacterial Index (CI), was initially developed by Wynne et al. \cite{wynne2008relating} and refined in \cite{wynne2010characterizing} to detect large, monospecific cyanobacteria blooms in Lake Erie. The CI exploits the unique pigment composition of cyanobacteria, particularly their phycocyanin accessory pigments and reduced chlorophyll‑\emph{a} fluorescence, which distinguishes them from eukaryotic phytoplankton. Chlorophyll absorption is calculated by spectral analysis around $681$ nm, leveraging top-of-atmosphere reflectance measured at specific wavelengths to determine the spectral shape and, subsequently, the CI value. However, limitations were encountered when estimating CI values, as highlighted first by the estimation of positive CI values without cyanobacteria measurements in Chesapeake Bay, Green Bay, and several New England lakes \cite{Wynne2018}. This was addressed by \cite{lunetta2015evaluation}, who introduced a conditional equation adjustment, resulting in the $\text{CI}_{Cyano}$ algorithm.  

The $\text{CI}_{Cyano}$ algorithm is widely validated for CyanoHABs monitoring in freshwater systems. Mishra \cite{mishra2021evaluation} validated $\text{CI}_{Cyano}$ for detecting toxin-producing cyanobacterial blooms across 30 U.S. lakes using MERIS and Sentinel-3 OLCI satellite data matched with field-measured microcystin concentrations. Using over 280 matchups from 2005-2019 and a microcystin threshold of $0.2\,\mu\text{g/L}$, the algorithm achieved 84\% accuracy with 90\% recall and 87\% precision when validated against both microcystin and cyanobacteria cell density. Coffer \cite{coffer2020quantifying} conducted a national-scale assessment of cyanobacterial blooms across 2,321 U.S. lakes, revealing strong seasonal and regional variability with typical summer-fall peaks. Wynne \cite{wynne2024remote} applied $\text{CI}_{Cyano}$ to monitor CyanoHAB variability in Lake Champlain's Missisquoi Bay, finding strong interannual variability linked to temperature and atmospheric instability. The algorithm has been operationally deployed across multiple U.S. states for bloom monitoring \cite{lunetta2015evaluation, clark2017satellite, stumpf2016challenges} and has informed state health advisories in California, Oregon, New York, Idaho, New Jersey, Utah, Vermont, and Wyoming \cite{schaeffer2018mobile, wyoming2018a, wyoming2018b, wyoming2018c}.

The CyAN employs the $\text{CI}_{Cyano}$ algorithm to estimate cyanobacterial concentrations within US lakes. CyAN is a collaborative effort between the United States Environmental Protection Agency, National Aeronautics and Space Administration, National Oceanic and Atmospheric Administration, and United States Geological Survey that leverages European Space Agency sensors: Medium Resolution Imaging Spectrometer (2002-2012) and Ocean and Land Color Instrument on Sentinel-3A/B (2016-present) to detect and quantify cyanobacterial algal blooms. Data from this effort are available as GeoTiff files containing either daily outputs or 7-day maximum value composites, both with a spatial resolution of 300m. We downloaded the daily GeoTiff files from January 2016 to November 2024 for our study. {\footnote{Dataset available at \url{https://oceancolor.gsfc.nasa.gov/about/projects/cyan/}} The broad time range provides sufficient temporal data to capture the dynamic nature of the CyanoHABs. 

\subsubsection{Temperature}
Previous studies \cite{gobler2017ocean, gobler2020climate, ho2020exploring, zhou2021effect, larsen2020extreme} suggested a strong correlation between temperature and CyanoHABs growth. JC Ho \cite{ho2020exploring} concluded that increased temperatures and longer summers increase chlorophyll concentrations, implying cyanobacterial growth. Therefore, we employed temperature data for CyanoHABs modeling.   

The temperature data are downloaded from the Moderate Resolution Imaging Spectroradiometer (MODIS) \cite{hulley2018viirs}, specifically from MODIS Land Surface Temperature and Emissivity. The data are in the form of GeoTiff files with a daily temporal resolution and spatial resolution of 1km and contain day and night temperatures.{\footnote{Dataset available at Google Earth Engine \url{https://developers.google.com/earth-engine/datasets/catalog/MODIS_061_MOD11A1}}

\subsection{Data Sparsity} \label{data_sparsity}
Remote sensing data for CyanoHAB detection face significant challenges due to missing values caused by cloud cover and sensor limitations, with approximately 35\% of the CI data and 87.88\% of the temperature data missing in Missisquoi Bay. Such sparsity disrupts the temporal continuity needed for effective time-series modeling. While sophisticated spatial reconstruction methods such as Data Interpolating Empirical Orthogonal Functions (DINEOF) \cite{beckers2003eof} have proven effective for satellite data gap-filling \cite{qiu2025gap, shi2023ocean, ye2025reconstructing}, we adopted a temporally-focused imputation strategy optimized for sequential forecasting. EOF-based approaches leverage spatial covariance but can attenuate onset/peak/decline transitions that are critical for learning CyanoHABs progression. Furthermore, our temperature dataset's extreme sparsity approaches the practical limits of iterative matrix decomposition, while our pixel-level approach with strict criteria (requiring $\geq 2$ valid observations within 3-day windows) and linear complexity supports efficient near-real-time operational forecasting. 

We present a two-stage imputation strategy to reconstruct bloom dynamics while preserving natural transitions between bloom states: (1) temporal data imputation and (2) temporal continuity restoration for CyanoHAB events. Imputation is performed on a segment-specific basis, ensuring that data from one segment does not affect others. Before processing, shallow pixels with a depth of less than 3m were filtered out using the official Lake Champlain bathymetry dataset provided by the Vermont Center for Geographic Information \cite{vermont_lake_champlain_bathymetry_2024}, as these pixels often display elevated CI values without actual blooms, allowing us to focus our analysis on pixels where CyanoHABs can be more reliably detected. 

\subsubsection{Temporal Data Imputation}
The imputation strategy utilizes two complementary approaches: Last Observation Carried Forward and Weighted Temporal Window Imputation, applied at the pixel level for both the CI and temperature datasets. While Weighted Temporal Window Imputation could be used alone, we first apply Last Observation Carried Forward to preserve the day-to-day progression of CyanoHABs dynamics better. This ordering ensures that when recent values suggest persistence, the imputation reflects that continuity rather than diluting intensity through averaging.

\begin{figure*}[!t]
\centering
\includegraphics[width=1\textwidth, height=1\textwidth, keepaspectratio]{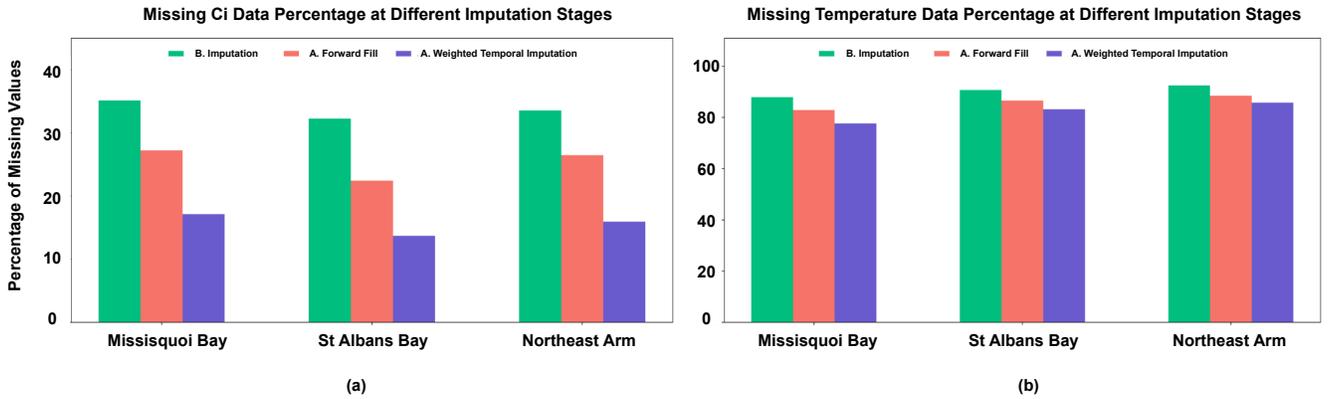}
\caption{Visualization of data sparsity. B. represents before while A. represents after.  (a) Shows Cyanobacterial Index Values missing percentage at different imputation stages, while (b) shows Temperature data missing percentage. The green bars illustrate around 30\% of the original Cyanobacterial Index Values and around 90\% of the Temperature data are missing. The orange bar shows the dataset after performing Forward Fill, indicating a significant addition of data points. The purple represents the dataset after Weighted Temporal Window Imputation, indicating further reduction in data missing percentage. We constrained further imputation to avoid adding too much noise.}
\label{data_imputation}
\end{figure*}

\textit{Last Observation Carried Forward (LOCF):} We implemented LOCF at the pixel level, examining each spatial coordinate sequentially across time. For pixels with missing values at time $t$, we checked if the exact spatial location contained a valid, non-imputed observation at time $t-1$. This value is propagated forward when available to fill the current missing value. This constraint maintains data integrity by balancing the need for completeness against the risk of introducing artificial patterns. LOCF is particularly important given the gradual day-to-day evolution typical in environmental systems. As shown in \Cref{data_imputation}, the LOCF application reduced the missing values of CI data at Missisquoi Bay from 35\% to 27\%, and missing temperature data from 87.88\% to 82.89\%.

\textit{Weighted Temporal Window Imputation:} After LOCF, we applied a weighted temporal window approach utilizing a 3-day historical window. For each missing pixel value at time $t$, we computed a weighted mean of the available observations over times $t-1, t-2$, and $t-3$, assigning more importance to recent observations. Specifically, we used weights of $3, 2$, and $1$ for $t-1, t-2$, and $t-3$, respectively. The imputed value $\hat{x}_{t}$ is calculated as:

\begin{equation} 
\hat{x}_t = \frac{3x_{t-1} + 2x_{t-2} + x_{t-3}}{3 + 2 + 1} 
\end{equation}

This imputation is conditionally applied only when at least two of the three observations within the window are valid (non-missing), ensuring sufficient information density for estimation. In cases where $t-1$ is missing, the imputation is based on the available values at $t-2$ and $t-3$, reducing the influence of recent trends but still capturing nearby temporal structure. We limited weighted temporal window imputation to a maximum of two consecutive days per pixel to prevent the propagation of error. As illustrated in \Cref{data_imputation}, weighted temporal imputation further decreased the percentage of missing CI data at Missisquoi Bay from 27\% to 17\%, while the percentage of missing temperature data values was reduced from 82.89\% to 77.69\%.

\subsubsection{Temporal Continuity Restoration in CyanoHABs Events}
Following data imputation, we observed discontinuities in CyanoHAB event signatures where gaps of 3-7 days created artificial interruptions in bloom progression. We implemented a targeted continuity restoration procedure specifically for CI data to address this limitation, as these values directly indicate the presence and intensity of CyanoHABs. Temporal discontinuities were systematically identified through visual inspection of daily CI data across all segments. When these discontinuities aligned with periods of missing data during active bloom phases, a simplified version of the weighted temporal imputation was applied. The missing pixel values were estimated using the weighted mean of valid observations from the previous three days, with all missing values excluded from the computation. All continuity restoration was performed at the pixel level before any spatial summarization, ensuring that temporal coherence is restored directly in the raw CI rasters rather than in the aggregated time series

The restoration preserved the natural temporal evolution of CyanoHABs events and assisted the model to learn accurate transition patterns between CyanoHABs states. Unlike the initial data imputation phase, which uniformly addressed all missing values, this restoration step was selectively applied only to positive CyanoHABs periods. Gaps longer than 7 days were excluded to avoid introducing excessive noise, and no restoration was performed for gaps between non-bloom days, as maintaining continuity in bloom-free intervals was deemed less critical for model performance.

\subsection{Data Extraction} \label{data_extraction}
\subsubsection{Cyanobacterial Index Data Extraction}
While forecasting the occurrence of CyanoHABs provides valuable information for policymakers and water resource managers, the practical utility of these forecasts increases substantially when intensity predictions are included. Traditional binary classification approaches predict simple presence or absence of CyanoHABs, while useful, fail to capture the critical nuance of bloom severity that drives management decisions. A low-intensity bloom may require only monitoring and public advisories, while an extreme bloom demands immediate interventions such as water treatment facility adjustments or recreational area closures. 

Motivated by this operational need, we formulated the CyanoHABs forecasting problem as a multi-class classification task to predict both occurrence and intensity levels, providing actionable intelligence about the expected severity range of CyanoHAB events. The CI values, which range from 0 to 253, directly correspond to cyanobacterial concentrations, where 0 indicates CyanoHABs absence and positive values indicate CyanoHABs presence, with larger values indicating higher bloom intensity. We established five distinct intensity categories to capture the full spectrum of bloom severity: Low Intensity, Medium Intensity, High Intensity, Very High Intensity, and Extreme Intensity CyanoHABs. The specific ranges for each intensity category are determined using equal-frequency binning applied separately to each segment's historical CI data. These categories allow stakeholders to anticipate the severity of upcoming events and implement proportional response measures. 

After establishing the five intensity categories, we developed a threshold detection mechanism to determine whether each category will be active (positive) or inactive (negative) for a given forecast period. A positive classification for any intensity category indicates that CyanoHABs events are expected to occur within that category's CI value range, enabling targeted management responses. For example, a positive prediction for the Extreme Intensity category signals the anticipated presence of a severe CyanoHAB event requiring immediate intervention, while a negative prediction suggests that extreme-level blooms are unlikely to develop during the forecast period.

To determine these thresholds, we implemented a data-driven approach that analyzes the historical distribution of pixel counts within each intensity category. Rather than applying arbitrary cutoffs, past CI values are examined, and for each day, the histogram of pixel counts is created for each intensity bin.  For Low and Medium intensity categories, we applied the 50th percentile as the threshold, requiring a substantial number of pixels to trigger a positive classification. This conservative approach is used to minimize false positives for these early forming CyanoHAB events. For the High-intensity category, we used the 60th percentile, acknowledging that while these events are more concerning, they still occur frequently enough to warrant moderate thresholds. However, for Very High and Extreme intensity categories, we adopted a highly sensitive approach with a threshold of one pixel. This ensures that even minimal evidence of severe bloom development triggers alerts, as missing these high-risk events could have serious public health and ecological consequences. Each segment's thresholds are calibrated independently using its specific historical data patterns.

To the best of our knowledge, no standardized ecological risk thresholds exist for CI-based CyanoHAB intensities. Therefore, we adopt a novel segment-specific binning strategy based on historical percentiles, enabling our model to reflect relative bloom severity and support actionable intensity-aware forecasting.

\subsubsection{Temperature Data Extraction}
Applying the same binning method used for CI data extraction might have resulted in unbalanced bin distributions due to missing temperature measurements and the dataset’s relatively coarser spatial resolution. Instead, we implemented a statistical approach to extract temperature statistics for each segment of interest and each day. We extracted five distinct statistical features from both daytime and nighttime temperature measurements: minimum, maximum, mean, standard deviation, and range. The range is the difference between the maximum and minimum temperature, while standard deviation provides the spread of the temperature from the mean temperature. This approach yields ten temperature-related features per day per segment, providing a richer representation of temperature conditions than simple binning would allow.

\subsection{Class Balancing}
Our dataset exhibits substantial class imbalance in target labels with a negative-to-positive ratio of approximately 5:1. This imbalance biases model predictions toward the majority class. To address this challenge, a targeted temporal sampling strategy is implemented. For each segment, months with consistent CyanoHABs formations are identified (e.g., June through October for Missisquoi Bay), and sequence creation is extended to include the month preceding these high-activity periods. Sequences and their multi-label targets are created using overlapping windows throughout this time frame (see section \ref{sequential_data_formation} for sequence and target creation), resulting in most samples representing positive CyanoHABs events. Negative samples from inactive or low-activity months (e.g., April, May, and November for Missisquoi Bay) are selected to balance the distribution. This approach preserves the critical temporal context needed for sequence modeling while achieving a more balanced class distribution. 

\section{Sequential Data Formation} \label{sequential_data_formation}
The dataset is structured to explicitly leverage its sequential nature by feeding the model temporally ordered input sequences, rather than isolated daily observations. In addition to the CI value and temperature data, temporal features such as the day of the year, season, and month are added using the timestamp data. The input sequence and targets are created segment-wise and then concatenated to have a single dataset to train the model on all segments combined. A rolling window approach is implemented for sequence and target creation, where each input sequence comprises 15 consecutive days with corresponding targets derived for a 14-day forecast horizon. The sequence window advances by one day for each new sample to ensure the capture of all temporal patterns. The segment signatures are omitted to force the model to learn a general CyanoHABs trend rather than segment-specific patterns. 

Given the daily data collection $D$ for each segment $s$ from a set of three monitoring segments $S$ - Missisquoi Bay, St. Albans Bay, and Northeast Arm. Each segment $s \in S$ contains data spanning $T_s$ days. Let $L=15$ represent the number of days used as input sequence length, and $H$ is the forecast horizon. 

For each starting day $t$ such that $1 \leq t \leq T_{s} - (L + H)$, we create an input sequence $X_{t}^{s}$ as:

\[X_t^s = \left(x_{t}^s, x_{t+1}^s, \ldots, x_{t+L}^s\right)\]

Here $X_t^s \in R^{L \times F}$ represents an input sequence for segment $s$ starting on day $t$, where $F$ represents the dimensionality of the daily feature vector. Using a rolling window approach, we advance one day forward for each new sequence, continuing until reaching the maximum valid starting day that allows for both a complete 15-day input sequence and the full $H$-day forecast horizon. 

For target creation, we classify each intensity bin for each day $h$ within the $H$-day forecast horizon. Let $b_{d, i}$ denote the pixel count in the $i^{th}$ intensity bin on day $d$. For each forecast day $h \in {1, 2, 3, \ldots, H}$ and each bin $i \in {1, 2, 3, 4, 5}$, the target is calculated as:

\begin{equation}
y_{t, h, i}^s = \begin{Bmatrix}
1 &  \text{if} \; b_{d=t+L+h, i}^s \geq \tau_{i}^{s}\\
0 & \text{Otherwise}  \\
\end{Bmatrix}
\end{equation}

where $\tau_i^s$ represents the threshold for $i^th$ bin for $s$. 

These thresholds are determined through statistical analysis of historical pixel distributions. For each segment and intensity bin, we extracted historical pixel counts and calculated the 50th percentile to establish thresholds $\tau = \left( \tau_1, \tau_2,\tau_3,\tau_4, \tau_5 \right)$. This approach ensures a balanced distribution between positive and negative classes rather than using arbitrary threshold values.

The target creation results in a multidimensional target tensor $Y_t^s \in {0, 1}^{H \times 5}$,  where each element indicates whether a specific intensity class will exceed its threshold on a particular forecast day. For a 14-day forecast horizon, this creates 70 binary classification targets per input sequence, providing detailed information about both the timing and intensity of potential bloom events.

\begin{figure*}[!t]
\centering
\includegraphics[width=1\textwidth, height=1\textwidth, keepaspectratio]{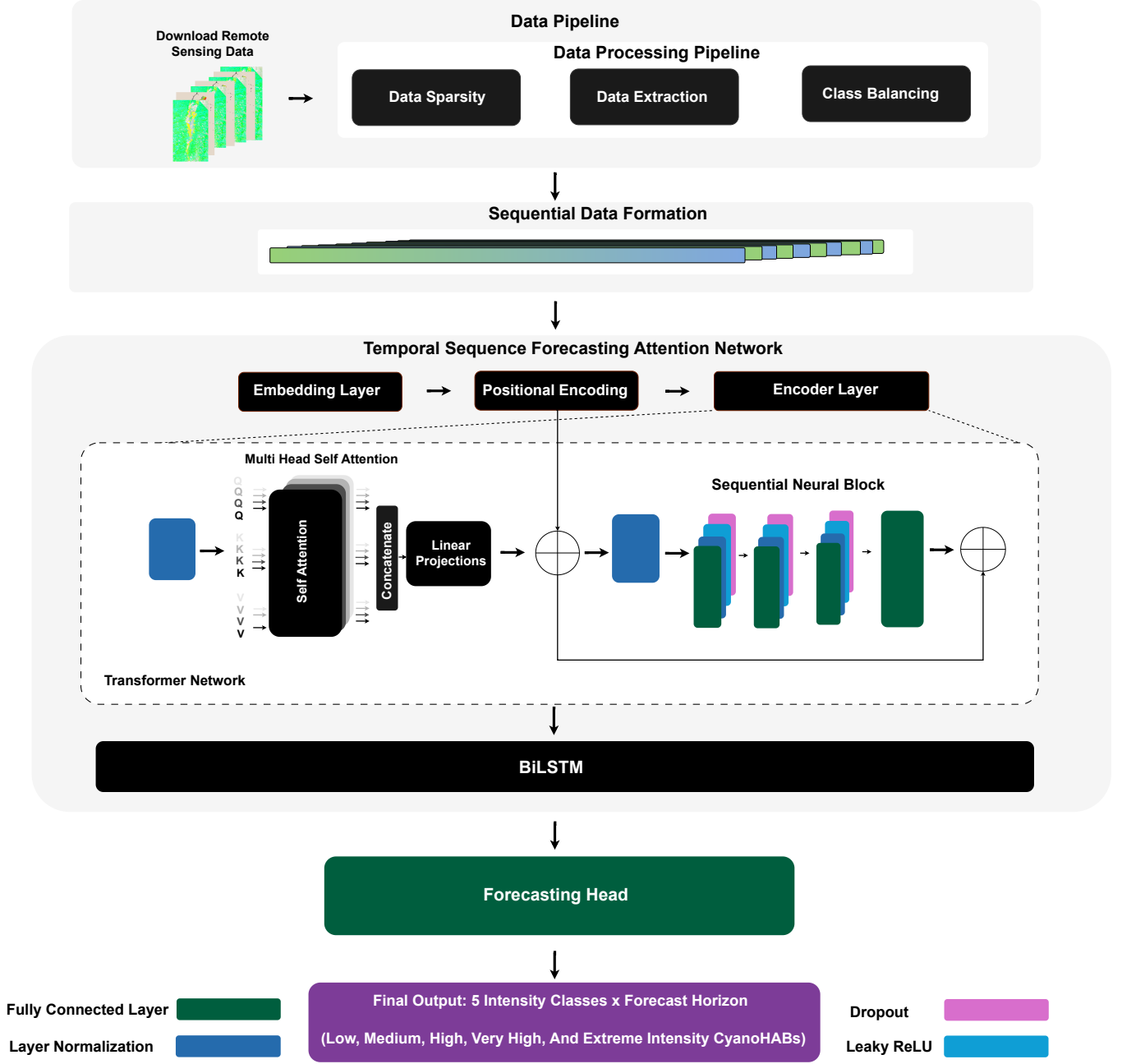}

\caption{The diagram illustrates the Transformer-BiLSTM model for CyanoHABs Intensity forecasting. The system processes 15 days of remote sensing inputs through an embedding layer, positional encoding, a Transformer encoder with multi-head self-attention, and a bidirectional LSTM block. The final output layer predicts five CyanoHAB intensity classes (Low to Extreme) across a 14-day forecast horizon. The pipeline incorporates preprocessing steps to address data sparsity and ensure class balance.}
\label{network_architecture}
\end{figure*}

\section{Methodology}\label{methodology}
Accurate CyanoHABs intensity forecasting is challenging due to the dynamics of CyanoHABs even within the same water body, and compounded by the impact of global warming affecting both the timing and spatial distribution of CyanoHABs. The proposed Transformer-BiLSTM incorporates a Temporal Sequence Forecasting Attention Network (TS-FAN), effectively capturing temporal patterns in sequential data features. TS-FAN combines an embedding layer to process the data sparsity, followed by sinusoidal positional encodings that provide temporal context. A multi-head self-attention block processes different
aspects of the sequence in parallel, while layer normalization and residual skip connections ensure stable training and improved gradient flow. A deep feed-forward network, structured as sequential neural blocks, refines feature representations through hierarchical transformations. A bidirectional LSTM further enhances sequential modeling by incorporating both past and future contexts. The forecasting task predicts the occurrence of future CyanoHABs events along with their intensities across five distinct levels (Low, Medium, High, Very High, and Extreme) for each day in a 14-day forecast horizon. Given an input sequence $X_{t}$, where $X_{t} \in \mathbb{R}^{L \times F}$ (see \Cref{sequential_data_formation}) with $L=15$ representing the sequence length and $F$ the number of features, we define an Transformer-BiLSTM function $f$ as:

\begin{equation}
Y_{t} = f(X_{t}) \quad \text{where} \quad f: \mathbb{R}^{L \times F} \rightarrow {0,1}^{H \times 5}
\end{equation}

The function $f$ maps the input time series with 15 time steps and $F$ features to an $H \times 5$ dimensional binary output matrix, where $H=14$ is the forecast horizon and 5 represents the intensity levels. For each forecast day $h \in {1, 2, ..., 14}$ and intensity level $i \in {1, 2, 3, 4, 5}$, the corresponding output element is either 1 to indicate the presence of CyanoHABs event within the specific intensity level or 0 to indicate the absence of the CyanoHABs event. This results in $70$ binary classification outputs per input sequence, providing detailed information about both the timing and intensity of potential CyanoHABs events. The overall network architecture is illustrated in \Cref{network_architecture}.

\subsection{Transformer-BiLSTM Architecture}
\subsubsection{TS-FAN}
Given the critical role of temporal dependencies in the CyanoHAB’s life cycle, it is essential to effectively capture the temporal relationships within the input sequence data. TS-FAN uses temporal feature extraction components to effectively model the complex temporal patterns and interactions inherent in the sequence data.

Given an input sequence $X_t$, it is first processed by an embedding layer. The layer takes the input sequence $X_t = \left ( x_{1},x_{2}, x_{3},..., x_{L} \right)$, where $L = 15$, each $x_t$ is $L \times F$ dimensional sparse feature vector, and it maps to a dense dimension $E_{n}$. Following \cite{vaswani2017attention}, we add positional encoding to the processed input sequence, allowing the model to understand the temporal structure. The positional encoding ($PE$) adds a unique vector at each position $p$ within a sequence and is defined. 

\begin{equation}
\begin{aligned}
PE(p, 2i) &= \sin\left( \frac{p}{10000^{2i/D'}} \right), \\
PE(p, 2i + 1) &= \cos\left( \frac{p}{10000^{2i/D'}} \right).
\end{aligned}
\label{eq:positional_encoding}
\end{equation}

Here, $p$ represents the timestamp position in the input sequence, and $i$ represents the position of dimension $D^{'}$, where $D^{'}$ is the dimensionality within the embedding space $E_n$. The division by $10000^{2_{i}/D^{'}}$ creates wavelengths that form a geometric progression from $2\pi$ to $10000 \cdot 2\pi$, allowing the model to learn to attend to relative positions with varying periods. This scaling ensures that the positional encodings for different dimensions have different frequencies, making each position uniquely identifiable.  Once the embeddings have been augmented with $PE$, the resultant vectors $E_{n}^{'}$ are forwarded to the encoder layer. We use a single encoder layer, which contains normalization layers, eight attention heads, and a sequential neural block. We replace the simple feed-forward network in the encoder layer with a deeper neural network, which we named sequential neural blocks. The input sequence $ E_{n}^{'}$ is passed to the encoder layer, where it is first processed by a layer normalization to stabilize the gradients. 

\begin{equation}
\begin{aligned}
E_{norm} = \text{LayerNorm}(E_{n}^{'})
\end{aligned}
\end{equation}

The normalized embeddings $E_{norm}$ are then processed by the multi-head attention mechanism within each head, where each head $h$ computes attention scores based on queries $Q$, keys $K$, and values $V$. Specifically, for each $h$, we have:\\

\begin{equation}
\begin{aligned}
Q_h,K_h, V_h = E_{norm}W_{h}^{Q},E_{norm}W_{h}^{K},E_{norm}W_{h}^{V}
\end{aligned}
\label{eq:head_attention_scores}
\end{equation}\\

Where, $W_{h}^{Q}, W_{h}^{K}, W_{h}^{V}$ are learnable weighted metrics of each head $h$, with assigned weights that are optimized during training. Here, the network takes the representation $Q_h$ and compares every other representation $K_h$ of the sub-sequence to determine the attention scores. Simply, it represents the relationship of each $Q_h$ with every other $K_h$, and the results are represented by values $V_h$. Subsequently, the softmax function is applied to normalize the attention scores between the range of 0 to 1, where 1 represents maximum correspondence, and it follows: \\

\begin{equation}
\begin{aligned}
Attention(Q_h, K_h, V_h) = softmax\left( \frac{Q_{h}K_{h}^{Transpose}}{\sqrt{d_{k}}}\right)V_{h}
\end{aligned}
\label{eq:head_attention_softmax}
\end{equation}\\

Here, $Transpose$ represents the transpose operation, and $d_k$ represents the dimensionality of the keys. The latter is used to scale down the dot product so that it does not become too large, facilitating stable gradient flow during training. 

Next, the outputs of all attention heads are concatenated and projected linearly before being added to the original input $E_{norm}$ via a residual connection to stabilize the learning process. The output of the multi-head self-attention, represented as $\vartheta$,  is passed to second layer normalization before being processed by the Sequential Neural Blocks. 

The multi-head self-attention module builds a complex understanding of temporal dependencies by processing sequential data patterns. To further refine these temporal representations, we replaced the traditional feed forward network in multi-head self attention with deeper and customized feed forward network named as Sequential Neural Blocks (SNB). SNB consists of three sequential blocks of fully connected layers, layer normalization, LeakyReLU, and a dropout layer, followed by a fully connected layer. These components refine complex interactions among features, stabilize learning, introduce non-linearity, and prevent overfitting. 

Given resultant output sequence $\vartheta$ from multi-head self-attention, it is passed to SNB, where it is processed by the first sequential block, mapping it to a new feature space:

\begin{equation}
\begin{aligned}
\zeta_{1} = \textup{Dropout}(\textup{L.ReLU}(\textup{L.Norm}(Z_1 = \vartheta W_{1} + b_{1})))
\end{aligned}
\label{eq:deep_neural_network_distribution_map}
\end{equation}

Here, $W_{1} \; \& \;  b_{1}$ are weight metrics and bias term of a fully connected layer of the first block, respectively, and  $Z_1$ represents the intermediate representation. This is passed to the layer norm to stabilize the $Z_1$ distribution, which is passed to the non-linear activation function to introduce the linearity. The subsequent dropout layer reduces the risk of overfitting. 

The output of the first block is represented as $\zeta_{1}$ and is passed to the second block to refine the abstractions further, which outputs $\zeta_{2}$. The third block further refines those representations and passes to the fully connected layer, which transforms the representations to a requisite feature dimension $\zeta_{4}$, ensuring compatibility for a residual skip connection. The $\zeta_{4}$ is added to the multi-head self-attention $\vartheta$ output before passing to the BiLSTM. This is shown below:

\begin{equation}
\begin{aligned}
En_{out} = \vartheta + \zeta_{4}
\end{aligned}
\end{equation}

The Encoder layer effectively captures temporal dependencies. We incorporate a BiLSTM layer to further enhance the model’s ability to process temporal information. It processes the encoded representations sequentially in both forward and backward directions, providing temporal context that strengthens the model’s predictive capabilities. Given $En_{out}$ as input from the encoder, BiLSTM computes the hidden states $\overrightarrow{h}$ and $\overleftarrow{h}$ in forward and backward directions, respectively. Afterward, the forward and backward outputs are concatenated to form a representation that encompasses information from past and future contexts.

\begin{equation} H_t =[\overrightarrow{h}; \overleftarrow{h}] \end{equation}

Here, $H_t$ represents the combined hidden state at the final timestamp $t$, and the semicolon denotes vector concatenation.

\subsubsection{Forecasting Head} 
The Forecasting Head consists of a fully connected layer and convert the model-specific temporal representation $H_t$ into a multi-output binary prediction over a 14-day forecast horizon. The output consists of 70 binary values representing the presence or absence of CyanoHABs across five intensity levels for each of the next 14 days. Formally, the forecasting head performs the following operation:

\begin{equation}
\text{Fout}_{t,h,i} = 
\begin{cases}
1 & \text{if } \sigma\left((H_t W_{fh} + b_{fh})_{h,i}\right) > \theta_{h,i},\\ 
0 & \text{otherwise}
\end{cases}
\end{equation}

\[\forall \, h \in \{1,\ldots,14\}, \quad i \in \{1,\ldots,5\}\]

Here, $\text{Fout}_{t,h,i}$ is the predicted binary label for day $h$ and intensity level $i$ for the input sequence at time $t$. The representation $H_t \in \mathbb{R}^d$ is the output of BiLSTM, and terms $W{fh} \in \mathbb{R}^{d \times 70}$ and $b_{fh} \in \mathbb{R}^{70}$ are the weights and bias of the fully connected output layer. The sigmoid function $\sigma$ maps each output to the range $[0, 1]$, and the threshold $\theta_{h,i}$ (typically set to 0.5) is used to determine the binary class assignment.

\subsubsection{Persistence Model (Baseline)}
To establish a comparison baseline, we include a Persistence Model that assumes bloom intensity classes remain unchanged over the forecast horizon. This approach is motivated by the temporal continuity observed in CyanoHAB events, where bloom conditions on day $t$ often persist to subsequent days. Formally, the persistence model assumes that the bloom intensity classes observed at time $t$ remain unchanged for each of the next 14 days:

\begin{equation}
\hat{y}_{t+h} = y_t, \quad \text{for} \quad h = 1, 2, \ldots, 14
\end{equation}

Here, $y_t \in {0,1}^5$ is the binary vector representing the presence or absence of CyanoHABs across five intensity levels (Low, Medium, High, Very High, Extreme) on day $t$. The prediction $\hat{y}_{t+h}$ at future horizon $h$ is identical to $y_t$, implying no evolution in bloom conditions.

While the persistence model does not involve any learning or parameter optimization, it serves as a strong short-term benchmark. Its relevance is especially notable in short-range forecasts (e.g., Day 1), where it occasionally outperforms learning-based models due to the slow-changing nature of bloom patterns. However, its inability to adapt to evolving environmental conditions leads to rapid performance degradation over longer horizons, thereby emphasizing the need for dynamic, data-driven approaches.

\subsection{Loss Functions}
The Transformer-BiLSTM outputs a binary prediction for the presence or absence of CyanoHABs across five intensity levels for each of the 14 forecast days, resulting in a total of 70 binary classification tasks per input sequence. To optimize model performance, we apply the Binary Cross-Entropy (BCE) loss to these outputs. The BCE loss is defined as:

\begin{equation}
\begin{aligned}
\text{BCE} \left( y_i, \hat{ y}_i\right) = -\frac{1}{N} \sum_{i=1}^{N} \left [ y_{i}\;log(\hat{y}_{i}) + (1 - y_{i})\;log(1 - \hat{y}_{i}) \right ]
\end{aligned}
\label{eq:bce_loss}
\end{equation}

Here, $y_i$ and $\hat{y}_i$ denote the ground truth and predicted probability for the $i^{th}$ binary output, respectively, and $N = 70$ is the total number of outputs per sequence. 

\section{Temporal Data Augmentation} \label{data_augmentation}
Transformer-based models typically require substantial training data to learn complex patterns and relationships effectively. However, the available dataset comprises only a few thousand sequences, which presents a significant challenge for model training. To address this limitation and enhance model generalization, the research implements a sophisticated temporal data augmentation strategy that respects the physical constraints and seasonality patterns of CyanoHABs.

The augmentation strategy is applied only during training to prevent data leakage into validation and test sets. It modifies two key aspects of the dataset: CyanoHABs intensity bins, and temperature statistics.

\subsubsection{CyanoHABs Intensity Bin Augmentation}
As introduced earlier (see \Cref{data_extraction}), the CI values were discretized into five segment-specific intensity categories - Low, Medium, High, Very High, and Extreme - using an equal-frequency binning strategy. Each of these categories corresponds to a CI intensity bin, representing a distinct range of bloom severity. In this section, we refer to these five intensity classes as bins for brevity.
Since CyanoHAB events exhibit strong seasonality, a month-dependent augmentation strategy introduces variability into bin values. Importantly, augmentation is applied only when a bin has at least three observed pixels, ensuring that artificial bloom events are not introduced during naturally inactive months. The magnitude and direction of adjustments vary seasonally: This is explained below:\\
\textit{Peak bloom months:}
\begin{itemize} 
 
\item If the bin count exceeds its segment-specific threshold by more than 10 pixels, it is considered a significant deviation. In this case, we randomly perturb the value by up to ±8 pixels to simulate realistic fluctuations in bloom intensity. 
\item If the count is above the threshold but with a smaller margin (1–10 pixels), we apply minor adjustments of up to ±3 pixels. 
\item If the count is below the threshold but above the minimum pixel threshold of 3, we apply small positive or negative changes up to ±3 to maintain variability near bloom onset. 

\end{itemize}

\textit{Non Peak bloom months:}
\begin{itemize} 
\item Adjustments are more conservative, with a maximum perturbation of ±3 pixels to reflect the naturally lower CyanoHABs activity.
\item If the bin count exceeds the threshold by more than 3 pixels, it is perturbed by up to ±3 pixels. 
\item If the count is just slightly above the threshold of minimum pixels of 3, we apply only positive perturbations of 0 to +2 to reflect very low cell counts. 

\end{itemize}

\subsubsection{Temperature Data Augmentation} 
Temperature serves a significant role in CyanoHAB development, where significant fluctuations can influence bloom dynamics. Controlled perturbations alter day and night temperature features while the system maintains physical consistency. This is explained below:
\begin{itemize} 

\item Day and night minimum and maximum temperatures are perturbed within a narrow range of –0.1°C to +0.16°C to simulate natural variability without altering ecological regimes. Larger perturbations (e.g., +5°C) could shift environmental conditions from moderate to warm, misleading the model. The selected range ensures stability of bloom-relevant conditions.  
\item Dependent features such as mean, range, and standard deviation are recalculated following perturbation to maintain internal consistency. 
\item Constraints enforce that day and night minimum temperatures remain lower than their respective maximum values to preserve physically valid relationships. 

\end{itemize}

This augmentation is applied to 50\% of the training sequences. The objective is to expose the model to plausible thermal variations while avoiding excessive artificial noise. Despite 90\% spatial missingness, the use of aggregated statistics across interpolated days provides a sufficient signal to benefit from this augmentation strategy.

\section{Results}
\subsection{Implementation Details} {\label{implementation_details}}
The Transformer-BiLSTM is implemented using PyTorch \cite{paszke2019pytorch}, a widely adopted deep learning framework. Several strategies are used to ensure robust training despite the challenges of sparse satellite data. 

For optimization, we employ the AdamW optimizer with a learning rate of 0.0005 and a weight decay of 0.09. The weight decay term helps reduce overfitting by discouraging excessively large weight values, promoting simpler and more generalizable models. A Cosine Annealing scheduler with Warm Restarts enhances the learning trajectory and cyclically resets the learning rate every five epochs. This approach allows the model to escape local minima and achieve better convergence. Early stopping prevents over-training by monitoring the validation's F1 performance and preserving the model's best state. Label smoothing \cite{goodfellow2016deep} enhances generalization by introducing a small amount of uncertainty in the target distribution, which prevents overconfidence in predictions. Gradient clipping with a threshold of 3.0 stabilizes training by preventing exploding gradients, which is particularly important for the recurrent components. The model is trained for 70 epochs with parameter updates governed by the minimization of BCE Loss. During evaluation, model outputs are thresholded using a default cutoff of 0.5. Any predicted confidence score $\hat{y} \geq 0.5$ or above is considered a positive prediction for the corresponding intensity class, while scores below this threshold are treated as negative.

The proposed model is trained on a personal computing setup equipped with a 13th Gen Intel Core i7-13700H processor, 32GB RAM, and an NVIDIA GeForce RTX 4070 GPU. This shows the model does not take much resources, which allows deployment in resource-constrained environments, making it practical for operational CyanoHAB intensity forecasting systems.

\subsection{Evaluation Metrics}
The forecasting model is evaluated using accuracy, precision, recall, and F1 scores. These are calculated as: \\
\[\small\text{Accuracy} = \frac{\text{TP + TN}}{\text{TP + TN + FP + FN} }\]
\[\small\text{Precision}    = \frac{\text{TP}}{\text{TP + FP}}\]
\[\small\text{Recall}    = \frac{\text{TP}}{\text{TP + FN}}\]
\[\small\text{F1 Scores} = \frac{2 \times \left (\text{Precision} \times \text{Recall}\right)}{\text{Precision + Recall}}\]\\

Here, TP, TN, FP, and FN are True Positives, True Negatives, False Positives, and False Negatives, respectively. In addition, we report the Area Under the Curve (AUC) to evaluate each model’s performance. AUC assesses the model’s ability to discriminate between classes across a range of classification thresholds.

\begin{table*}[t]
\centering
\setlength{\tabcolsep}{6pt}
\renewcommand{\arraystretch}{1.4}
\caption{Performance metrics for individual models and the Transformer-BiLSTM for the 14th day across a 14-day forecast horizon. Transformer-C. Attention represents the Transformer with a Custom Attention Layer. The Transformer-BiLSTM consistently outperformed all models across all metrics for long-term CyanoHABs forecasting.}
\label{tab:overall_results}
\begin{tabular}{
    >{\centering\arraybackslash}p{2.2cm}
    >{\raggedright\arraybackslash}p{2.8cm}
    >{\centering\arraybackslash}p{1.8cm}
    >{\centering\arraybackslash}p{2.2cm}
    >{\centering\arraybackslash}p{2.0cm}
    >{\centering\arraybackslash}p{1.8cm}
    >{\centering\arraybackslash}p{1.8cm}
}
\hline
\rowcolor[gray]{0.95}
\textbf{Model Category} & \textbf{Model} & \textbf{Acc (\%)} & \textbf{Precision (\%)} & \textbf{Recall (\%)} & \textbf{F1 (\%)} & \textbf{AUC (\%)} \\
\hline

\multirow{1}{*}{\textbf{Baseline}} 
    & Persistence Model & $71.34$ & $71.68$ & $71.34$ & $71.54$ & $70.42$ \\
\arrayrulecolor{gray!50}\hline

\multirow{3}{*}{\textbf{Machine Learning}} 
    & Support Vector Machines & $43.29$ & $62.20$ & $51.91$ & $55.29$ & $80.12$ \\
    & XGBoost & $41.77$ & $61.22$ & $53.67$ & $56.68$ & $80.16$ \\
    & Random Forest & $45.45$ & $63.38$ & $57.89$ & $59.36$ & $82.11$ \\
\arrayrulecolor{black}\hline

\multirow{7}{*}{\textbf{Deep Learning}} 
    & Transformer & $74.95$ & $74.35$ & $74.95$ & $74.08$ & $77.18$ \\
    & Transformer-C.Attention & $75.08$ & $74.48$ & $75.08$ & $74.30$ & $77.04$ \\
    & Transformer-BiGRU & $77.13$ & $76.97$ & $77.13$ & $77.04$ & $81.01$ \\
    & Transformer-LSTM & $77.47$ & $77.39$ & $77.47$ & $77.43$ & $80.85$ \\
    & BiLSTM & $77.71$ & $77.38$ & $77.71$ & $77.46$ & $80.32$ \\
    & Transformer-GRU & $77.95$ & $78.43$ & $77.95$ & $78.13$ & $82.53$ \\

\rowcolor[gray]{0.95}
    & \textbf{Transformer-BiLSTM} & \textbf{78.94} & \textbf{78.80} & \textbf{78.94} & \textbf{78.86} & \textbf{82.55} \\
\hline
\end{tabular}
\end{table*}

\begin{figure*}[t]
\centering
\includegraphics[width=1\textwidth, keepaspectratio]{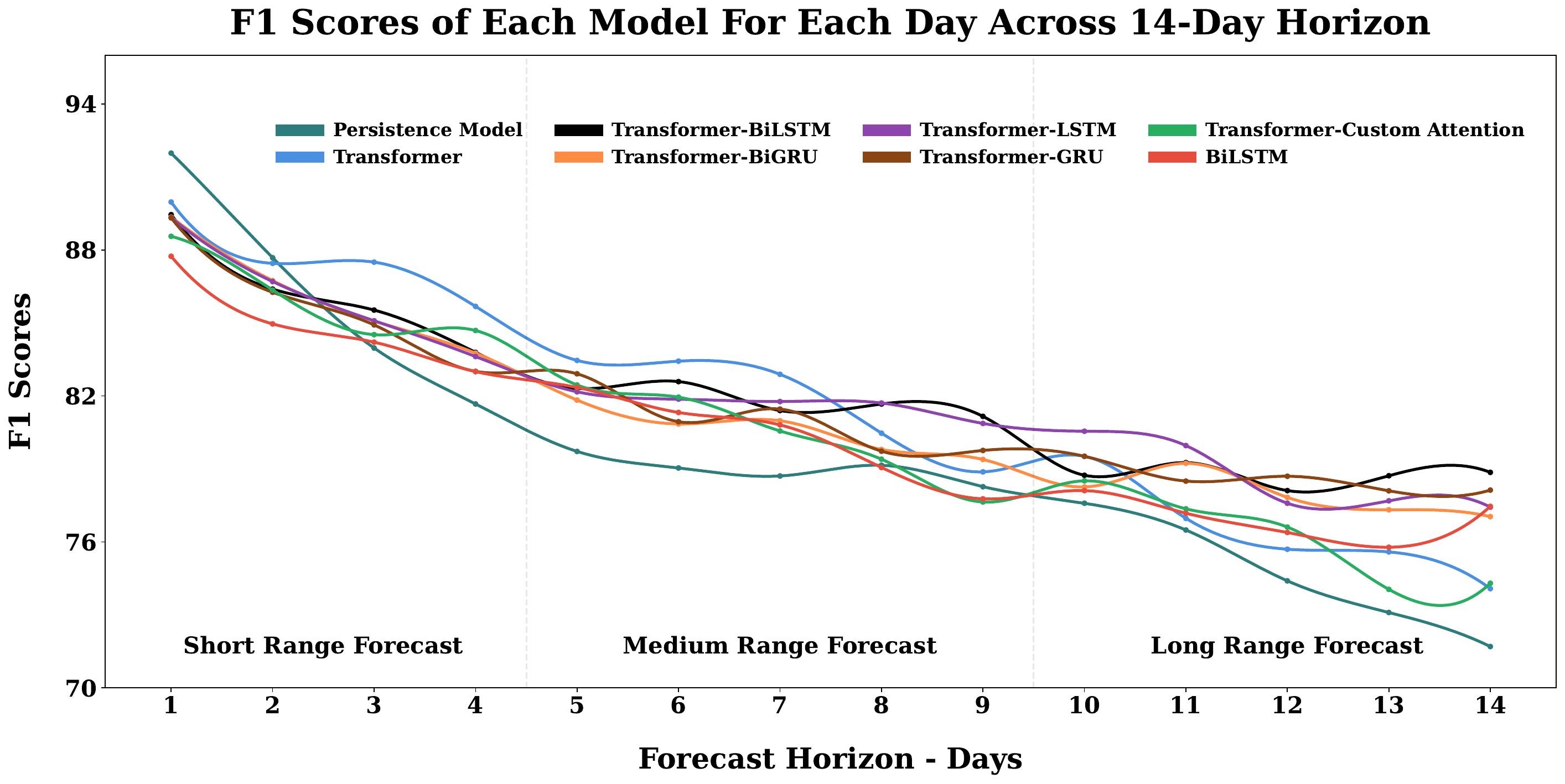}
\caption{Visualization F1 scores of each model for each day across the 14-day forecast horizon for all segments. F1 score trends across a 14-day forecast horizon for all individual models and the Transformer-BiLSTM model. Forecast performance generally declines with increasing lead time. The Transformer-BiLSTM outperforms individual model across longer-forecast ranges and perform competitively for short and medium forecast. Vertical dashed lines demarcate short-range (Days 1–4), medium-range (Days 5–9), and long-range (Days 10–14) forecasts. Circular markers denote discrete evaluation points at each forecast day.}
\label{f1_scores_analysis_across_14_days}
\end{figure*}

\subsection{Intensity Forecasting Evaluation} \label{intensity_evaluation}

\subsubsection{Quantitative Evaluation} \label{forecasting_quantitative_evaluation}
To evaluate the proposed Transformer-BiLSTM for CyanoHAB intensity forecasting, we compared it with two sets of architectures designed explicitly for this study. The first set covers three commonly used machine-learning algorithms, which include Random Forest (RF), Support Vector Machine (SVM), and XGBoost. The second set consists of six custom deep-learning architectures. These include standalone Transformer-only and BiLSTM-only models, as well as four hybrid variants that pair a Transformer encoder with LSTM, BiLSTM, GRU, or BiGRU modules, plus a version that replaces the standard attention with a Custom-Attention block. This benchmark provides a comprehensive performance evaluation against both traditional and state-of-the-art approaches.

For standalone architectures, the Transformer-only model processes input directly through the encoder layer to the forecasting head, while the BiLSTM-only model employs an embedding layer before BiLSTM processing and subsequent prediction. The hybrid configurations combine transformers with Long Short Term Memory (LSTM) \cite{hochreiter1997long}, Gated Recurrent Unit (GRU) \cite{cho2014learning}, and Bidirectional Gated Recurrent Unit (BiGRU), alongside a Custom Attention layer. In these hybrid models, encoder outputs serve as inputs to their respective recurrent or attention blocks, e.g., Transformer-BiGRU forwards the encoded sequence to a BiGRU layer for bidirectional processing. Besides utilizing established recurrent units, we incorporated a custom attention layer to provide additional emphasis on the most important time steps essential for accurate prediction. The intuition behind this approach is to assign learnable weights to the encoder layer outputs from the transformer network, thereby enhancing the model's focus on temporally critical features further. The Transformer-Custom Attention variant computes context-aware summaries by assigning learnable scalar weights to each time step, which are normalized via softmax function and used to compute weighted sums of encoder outputs, thereby emphasizing temporally important features before forecasting.

All deep learning based models were trained under identical conditions as detailed in Section~\ref{implementation_details}, using data from 2016 to 2021 for training, 2022 for validation, and 2023-2024 for testing, maintaining consistent data preprocessing pipelines, loss functions, and optimizer configurations with the proposed Transformer-BiLSTM architecture. Training was standardized at 70 epochs across all models, with the exception of the custom attention variant, which required extended training of 120 epochs due to its additional learnable parameters and increased computational complexity. Performance evaluation utilized standard metrics, including accuracy, precision, recall, and F1 score.

The machine learning models were trained on identical data splits but required different preprocessing since Random Forest, XGBoost, and SVM operate on fixed-length vectors rather than sequences. Input data was flattened before processing. The Random Forest comprised 600 trees grown to full depth with a minimum of two samples per leaf, wrapped in a MultiOutputClassifier from Scikit-learn \cite{scikit-learn} to predict 70 binary labels simultaneously (14 days × 5 classes). The SVM used RBF-kernel with probability estimates enabled, embedded in a One-Vs-Rest wrapper from Scikit-learn to train one classifier per label for consistent AUC computation. XGBoost followed the same preprocessing pipeline and One-Vs-Rest ensemble approach, learning one boosted tree model per output label.

\Cref{tab:overall_results} summarizes the forecasting performance of each model over a 14-day prediction horizon. Notably, the metrics presented in \Cref{tab:overall_results} represent the performance specifically for the 14th day forecast. The results show deep learning models significantly outperform traditional machine learning approaches. 

The Transformer-BiLSTM architecture achieves the highest performance across all metrics, with 78.94\% accuracy, 78.80\% precision, 78.94\% recall, 78.86\% F1 score, and an AUC of 82.65\%. This highlights the effectiveness of combining self-attention mechanisms with bidirectional sequence modeling to capture both contextual information and temporal dependencies in CyanoHAB dynamics. The Transformer-GRU and standalone BiLSTM models also performed strongly, achieving F1 scores of 78.13\% and 77.46\%, and AUC of 82.53\% and 80.32\%, respectively. Notably, all hybrid architectures consistently outperformed the baseline Transformer model over a more extended forecast period. While the Transformer-Custom Attention Layer showed modest improvement over the baseline, the integration of recurrent components yielded substantially better results. This highlights the importance of integrating both attention-based architectures and temporal modeling for extended forecasting tasks, as neither mechanism alone has proven sufficient to maintain performance at longer prediction intervals. 

The machine-learning models trail deep learning based models in point metrics; however, their AUC scores are comparatively high. Random Forest reaches an AUC of 82.1 \%, and both SVM and XGBoost exceed 80 \%, despite accuracy in the low to mid 40\% range. This disparity indicates that tree and kernel-based methods rank bloom risk reasonably well but struggle to convert those rankings into correct hard labels once a decision threshold is set. In other words, they discern separability yet lack calibrated decision boundaries for the multi-label, highly imbalanced setting

The Persistence Model, used as a naive temporal baseline for deep learning based models, achieved an F1 score of 71.70\%. While it performed competitively in short-term forecasts (see \Cref{f1_scores_analysis_across_14_days}), its performance significantly deteriorated over longer horizons, as exemplified by its last-place ranking on Day 14. This trend reflects its inability to adapt to dynamic bloom transitions and further underscores the value of learning-based models for medium to long-range forecasting.

One can observe that Transformer-BiLSTM consistently delivered the most balanced performance across all metrics. This suggests that the combination of bidirectional sequential modeling and self-attention leads to more temporally coherent and context-rich representations. The Transformer's global attention captures long-range dependencies while BiLSTM's dual-directional processing models local temporal patterns, enabling the architecture to effectively identify subtle CyanoHAB dynamics such as bloom progression and decay phases. This combined approach proves particularly valuable for extended forecast horizons, where neither attention mechanisms nor recurrent processing alone maintains optimal prediction accuracy.\

Further F1 scores analysis was conducted to examine how model performance changes across the forecast horizon. Since the Machine Learning based model showed lower F1 scores, this comparison focused on the deep-learning architectures and the Persistence model only. \Cref{f1_scores_analysis_across_14_days} illustrates the F1 scores of the persistence model and each deep learning model for each day across the 14-day forecasting period. In the short-range forecasts (Days 1–4), all models demonstrate strong performance, with the standalone Transformer performing comparably to the Transformer-BiLSTM during the initial days. The Persistence Model, due to the temporal continuity of CyanoHAB events, performs competitively on Day 1, even marginally outperforming learning-based models. However, this advantage is short-lived; its static assumption becomes increasingly inaccurate with time, and it quickly deteriorates, reaching the lowest F1 score by Day 14. 

A notable transition occurs as the forecast horizon extends into the long-range period (Days 10-14). While the standalone Transformer shows a steep decline in performance beyond Day 9, the Transformer-BiLSTM model takes control and consistently outperforms all other models in the long-range period. This superior long-range performance reflects the hybrid architecture's ability to effectively combine short-term contextual understanding via attention mechanisms with robust long-term dependency modeling through bidirectional recurrence. Other hybrid models like Transformer-GRU also show better stability than the standalone Transformer, but the Transformer-BiLSTM consistently achieves the highest F1 scores in long-range CyanoHAB intensity forecasting.

\begin{figure*}[h]
\centering
\includegraphics[width=1\textwidth, keepaspectratio]{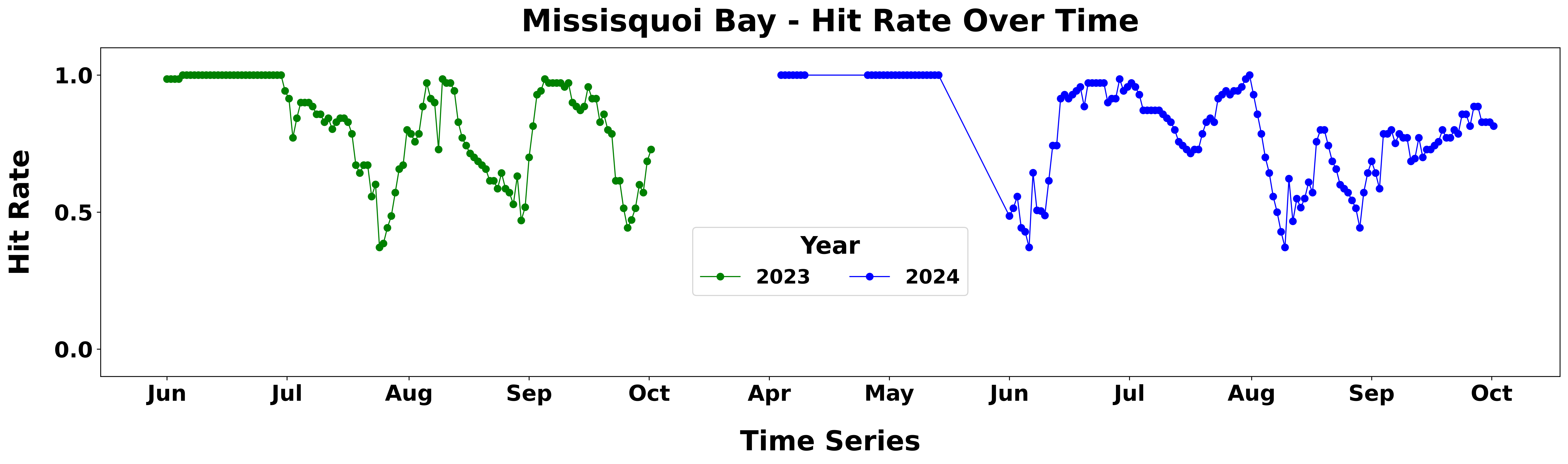}
\caption{Temporal hit rate analysis for Missisquoi Bay across the 2023-2024 test period. The visualization reveals seasonal patterns in CyanoHAB prediction accuracy, with higher hit rates during established CyanoHAB phases and variation during transitions. The patterns between 2023 and 2024 demonstrate the model's adaptability to different CyanoHAB dynamics while maintaining robust seasonal forecasting capabilities.}
\label{mq_hitrate_results_20232024}
\end{figure*}
\begin{figure*}[h]
\centering
\includegraphics[width=1\textwidth, keepaspectratio]{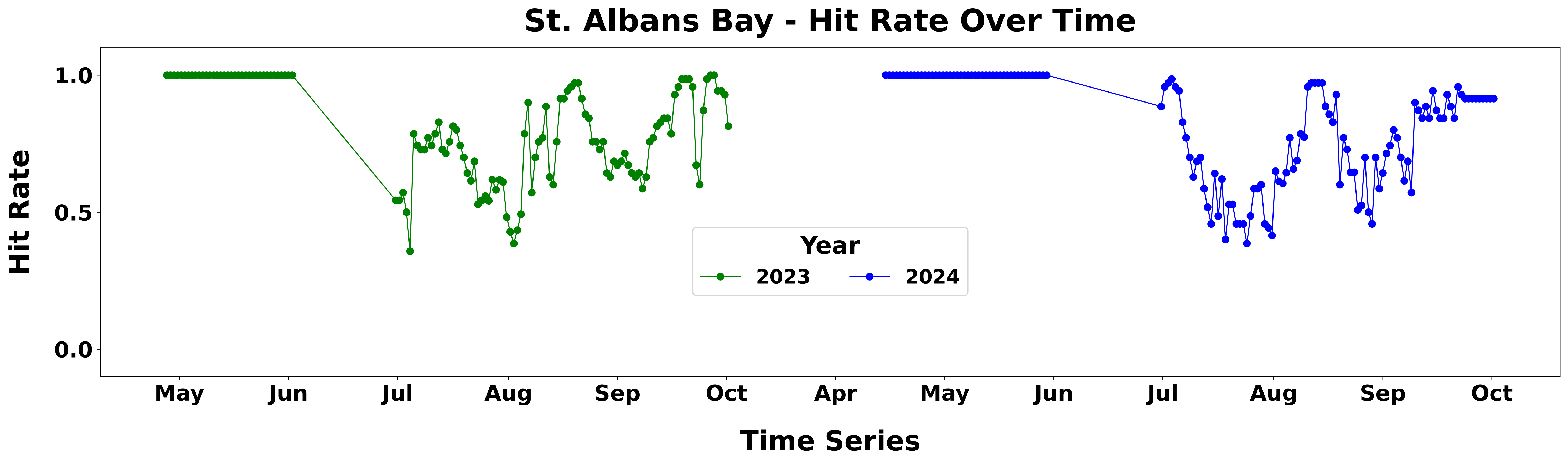}
\caption{Temporal hit rate analysis for St. Albans Bay across the 2023-2024 test period. The visualization shows later CyanoHABs formations compared to Missisquoi Bay. The fragmented CyanoHABs patterns create more frequent hit rate fluctuations during the season.}
\label{st_hitrate_results_20232024}
\end{figure*}

\subsubsection{Qualitative Evaluation}
To evaluate the qualitative performance of our forecasting models, we employed the Probability of Detection (POD), also known as the Hit Rate. POD quantifies the proportion of observed events that are correctly predicted by the model, providing insight into the model's ability to identify true positive occurrences.

The use of POD in harmful algal bloom (HAB) forecasting has been well-documented in previous studies. For instance, the National Oceanic and Atmospheric Administration (NOAA) utilized POD \cite{kavanaugh2013assessment} to assess the performance of its HAB Operational Forecast System in the western Gulf of Mexico. This system provided forecasts of bloom transport and associated respiratory irritation, with POD serving as a key metric for evaluating forecast accuracy over multiple years. Similarly, the California Harmful Algae Risk Mapping system was designed to predict the spatial likelihood of Pseudo-nitzschia blooms and associated domoic acid events along the California coast \cite{anderson2016initial}. In its initial skill assessments, the system incorporated POD by comparing model-generated probabilities with observed bloom occurrences to evaluate forecasting reliability.

Mathematically, the Probability of Detection is defined as:

\[\small\text{POD/Hit Rate} = \frac{\alpha}{\alpha + c }\]

Here, $\alpha$ represents the number of hits (instances where an event was both observed and correctly forecasted) and $c$ denotes the number of misses (instances where an event was observed but not forecasted).

The POD value ranges from 0 to 1, with a value of 1 indicating perfect detection (all observed events were correctly forecasted), and a value of 0 indicating no detection (none of the observed events were forecasted). Intermediate values reflect the proportion of observed events that were successfully predicted.

While POD is formally equivalent to Recall, we used it as a concise visual summary to interpret multihorizon predictions. Given the 14-day forecast window, directly comparing daily predictions can be visually cluttered. The POD thus provides a readable behavior of detection consistency across the forecast horizon, supporting qualitative model assessment. We calculated the POD for each station-day by comparing the model's predictions with the actual observations. The comprehensive analysis for each station's hit rate is presented below.

\begin{figure*}[h]
\centering
\includegraphics[width=1\textwidth, keepaspectratio]{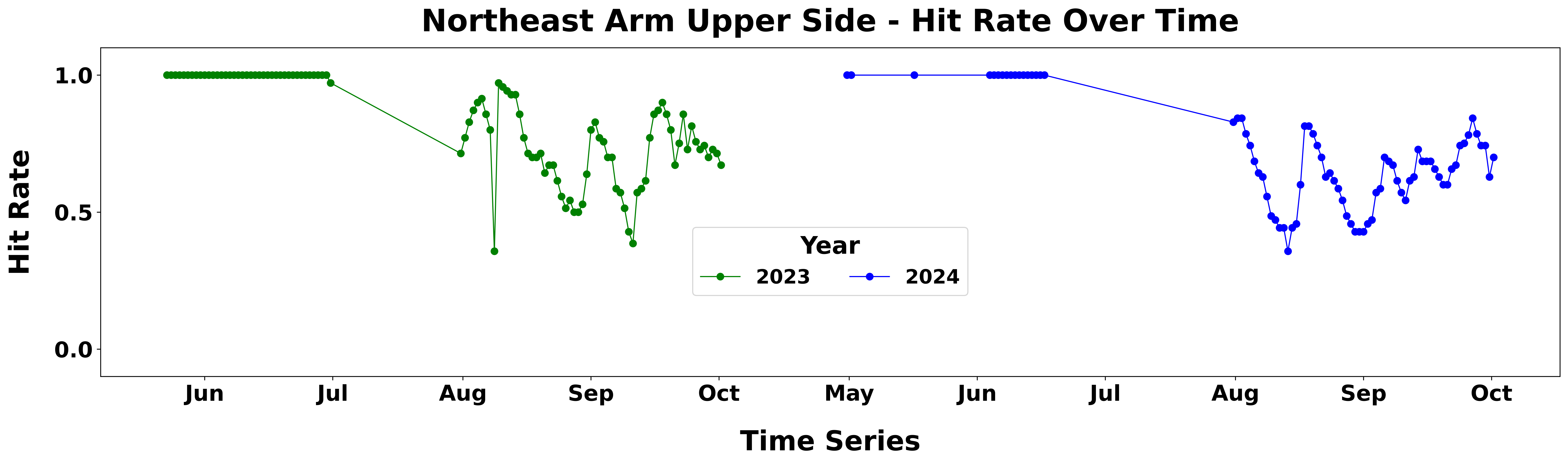}
\caption{Hit rate time series for the Northeast Arm Upper Side segment across 2023 and 2024. The diagram illustrates consistent fluctuations in hit rate during August to October, corresponding to fragmented and short-lived CyanoHAB events.}
\label{ne_upper_hitrate_results_20232024}
\end{figure*}

\begin{figure*}[h]
\centering
\includegraphics[width=1\textwidth, keepaspectratio]{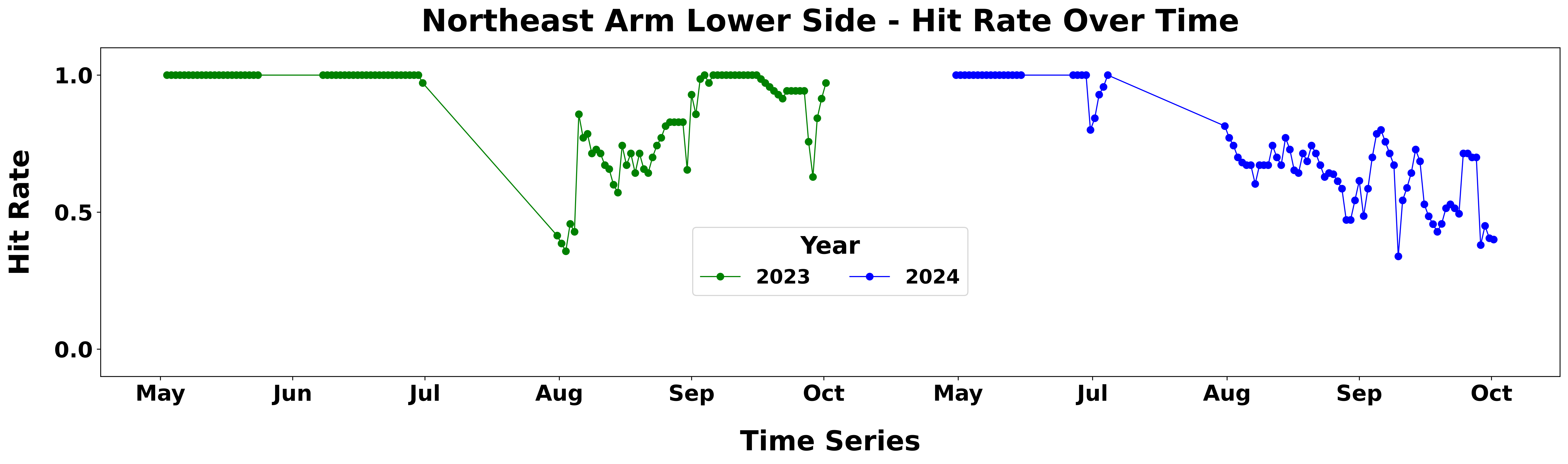}
\caption{Hit rate time series for the Northeast Arm Lower Side segment across 2023 and 2024. In 2023, the model achieves stable performance during sustained CyanoHABs periods in late September, reflecting more persistent CyanoHAB activity. In contrast, 2024 exhibits increased variability in hit rates, indicating fragmented and intermittent CyanoHABs patterns.}
\label{ne_lower_hitrate_results_20232024}
\end{figure*}

\paragraph{\textbf{Missisquoi Bay}} \label{mq_results_visualization}
The time series visualization for Missisquoi Bay (\Cref{mq_hitrate_results_20232024}) illustrates the hit rate analysis during 2023 and 2024. The visualization demonstrates the model's ability to capture CyanoHAB seasonal patterns across different CyanoHABs phases. During non-CyanoHAB periods, the model achieves a higher hit rate, as seen in June 2023, accurately identifying the absence of CyanoHABs across all intensity categories. During transition periods when CyanoHABs begin forming or dissipating, hit rates naturally decrease, reflecting the inherent challenge of predicting these dynamic phases. This pattern is visible in early August 2023, where the hit rate transitions from 1.0 to approximately 0.5. The model maintains strong performance during established CyanoHABs periods, with hit rates stabilizing between 0.8 and 1.0, demonstrating its capability to forecast stable CyanoHAB states accurately.

The comparison between 2023 and 2024 reveals the model's adaptability to varying CyanoHAB dynamics. The 2023 pattern shows a clear seasonal progression: non-CyanoHABs period (June), transition phase (July), peak CyanoHABs (August-September), and gradual decline (October). The 2024 data presents a more complex pattern with earlier CyanoHABs initiation in June and multiple formation-dissipation cycles throughout the season. Despite these variations, the model successfully tracks the overall seasonal trends in both years, with hit rates recovering during peak CyanoHABs periods and maintaining high accuracy during stable conditions.

These temporal patterns illustrate the model's robust performance in capturing CyanoHAB seasonal dynamics. While hit rates fluctuate during transition periods - a natural characteristic of environmental forecasting - the model consistently identifies the broader seasonal patterns and maintains high accuracy during both CyanoHABs and non-CyanoHABs stable states. 

\begin{figure*}[h]
\centering
\includegraphics[width=1\textwidth, keepaspectratio]{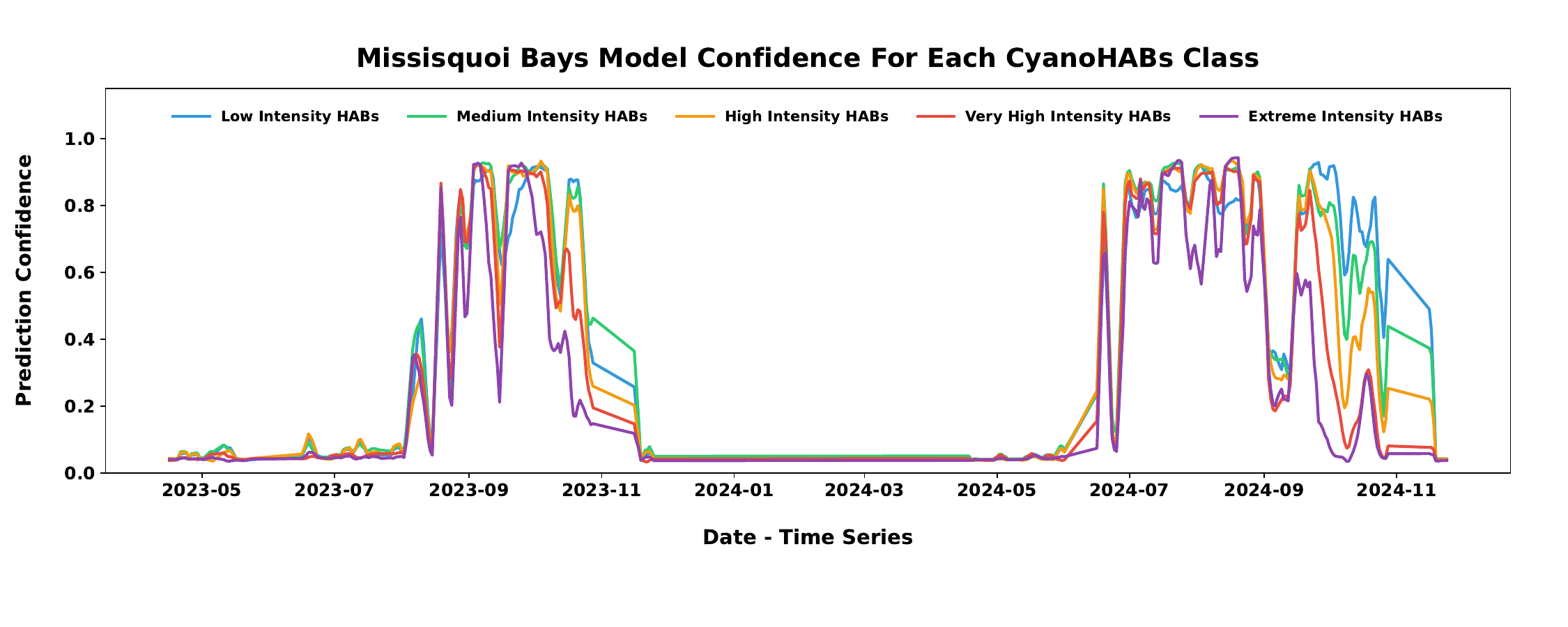}
\caption{Model prediction confidence scores for five CyanoHAB intensity classes in Missisquoi Bay. The graph displays confidence values for Low (blue), Medium (green), High (orange), Very High (red), and Extreme (purple) intensity classes. High confidence scores align with known bloom periods in August to October 2023 and June to September 2024, while near-zero confidence appears during non-bloom seasons, demonstrating the model's ability to recognize seasonal patterns. }
\label{mq_confidence_scores}
\end{figure*}

\begin{figure}[h]
\centering
\includegraphics[width=1\columnwidth, height=1\columnwidth, keepaspectratio]{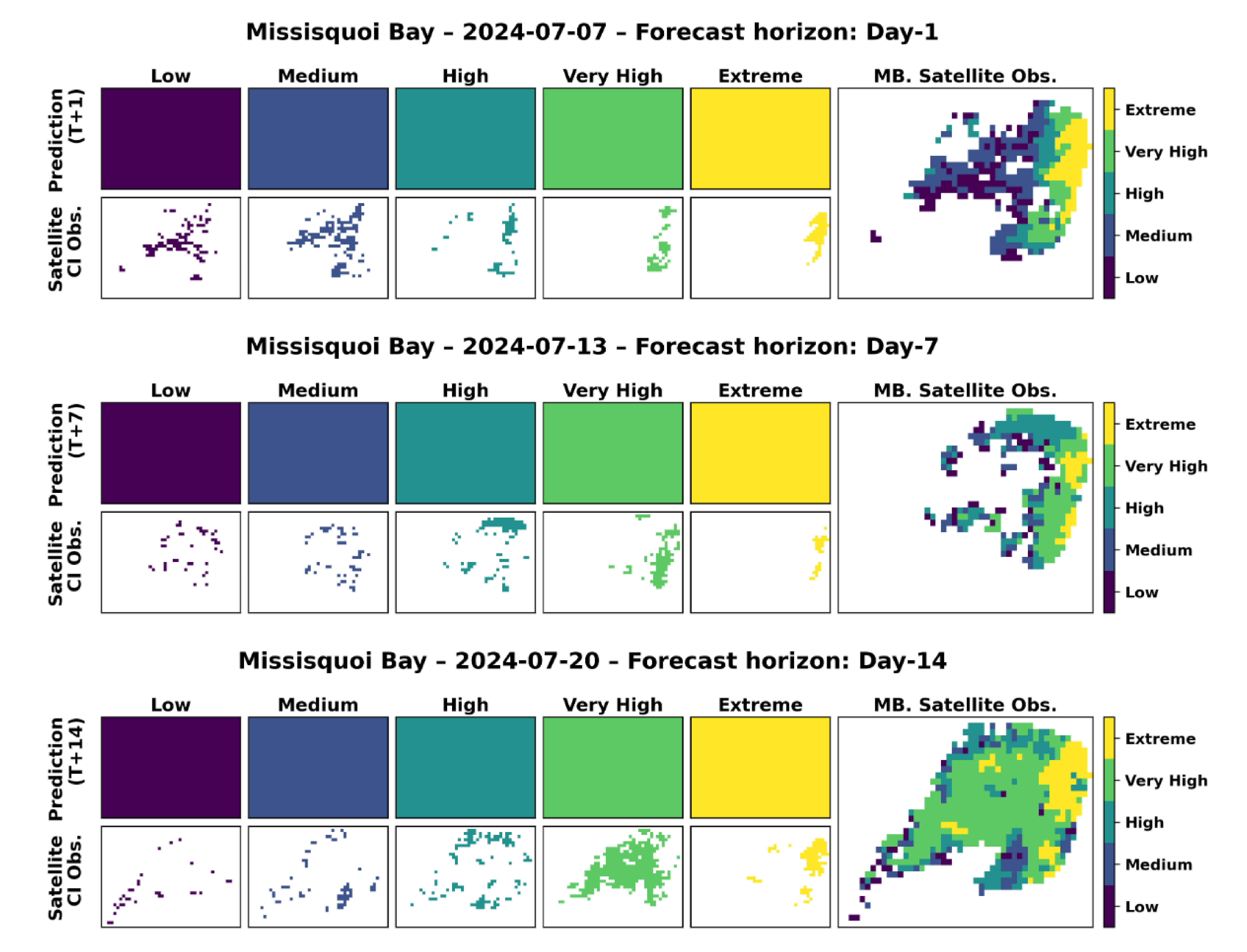}
\caption{Spatial comparison between model forecasts and CyAN satellite observations of Missisquoi Bay during July 2024. Each row gives a different lead time—Day-1 (7 July), Day-7 (13 July), and Day-14 (20 July). The top shows the Transformer-BiLSTM prediction for each CI intensity class (Low, Medium, High, Very High, Extreme); the bottom shows the corresponding satellite mask, followed by the composite satellite map on the far right. One can observe the missing pixels in the complete composite of the Missisquoi Bay station. The model correctly preserves the correct class mix over the full 14-day horizon.}
\label{mb_maps_1}
\end{figure}

\begin{figure}[h]
\centering
\includegraphics[width=1\columnwidth, keepaspectratio]{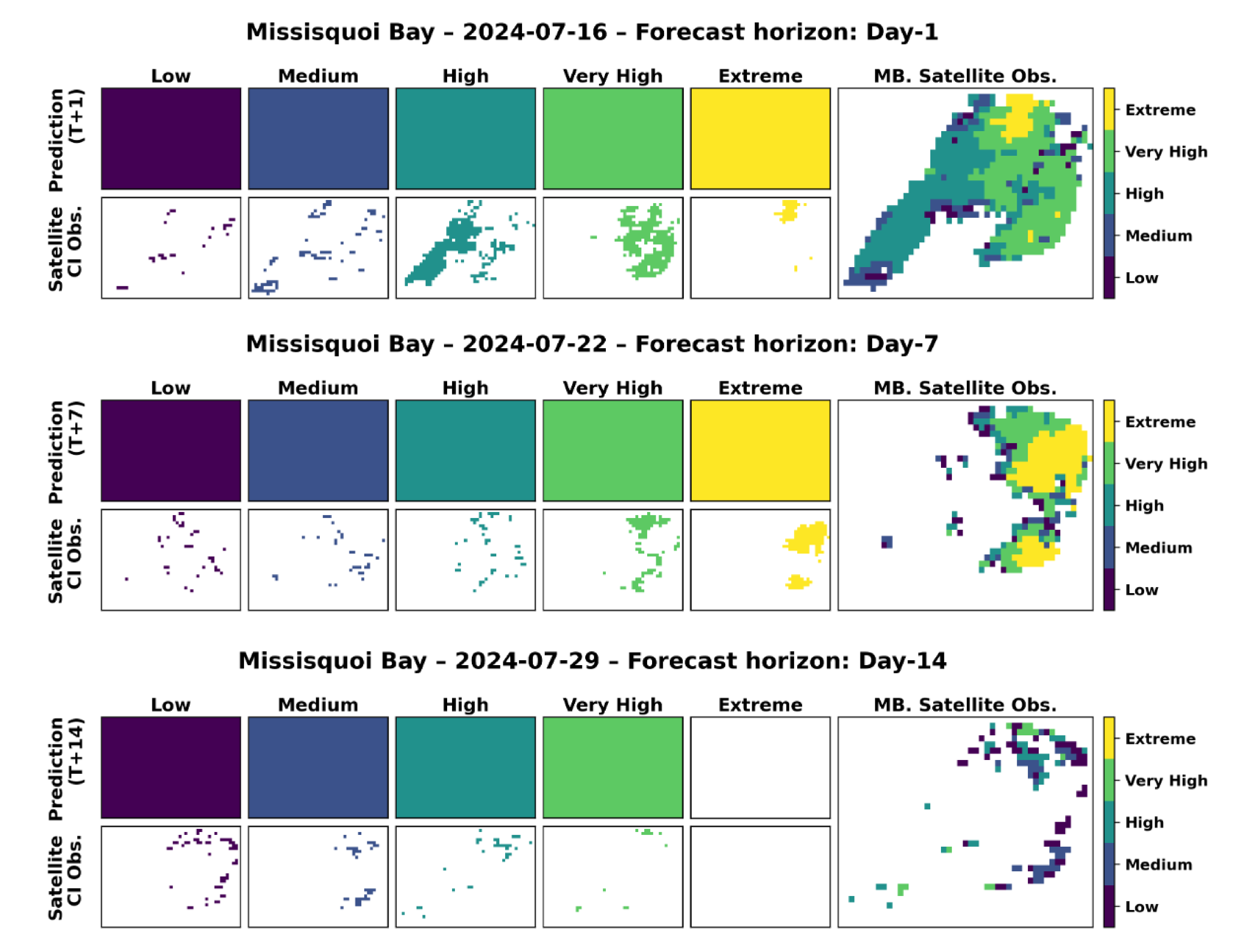}
\caption{Spatial comparison between model forecasts and Cyan satellite observations of Missisquoi Bay from 16 July to 29 July, 2024. One can observe the alignment between predictions and targets. We can also see that on July 29, the targeted image is sparser compared to Day-7 and Day-1 images, which adds an additional layer of complexity due to not having enough pixels for predictions.  }
\label{mb_maps_2}
\end{figure}

\begin{figure}[h]
\centering
\includegraphics[width=1\columnwidth, keepaspectratio]{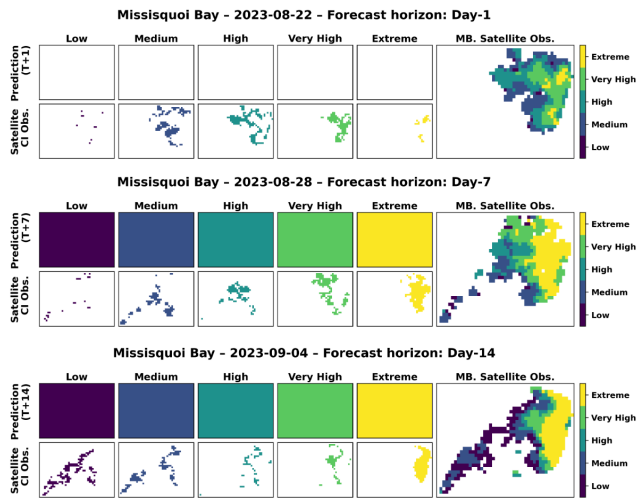}
\caption{Same layout as earlier diagrams. The Day-1 forecast underestimates the initial outbreak, yet the Day-7 (28 August) and Day-14 (4 September) predictions catch up to the observed magnitude and intensity distribution. This rapid convergence shows that while abrupt bloom onsets can be missed, the model quickly adjusts once the larger-scale environmental drivers dominate. }
\label{model_underpredictions}
\end{figure}

\begin{figure}[h]
\centering
\includegraphics[width=1\columnwidth, keepaspectratio]{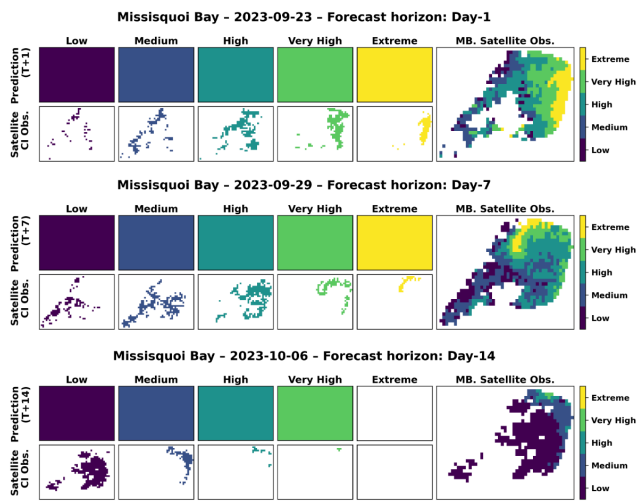}
\caption{Forecast skill during a late-season bloom that extends from 23 September to 10 October 2023 in Missisquoi Bay. Both the model and the observations maintain High and Extreme-class pixels well past the usual summer window, confirming that the bloom persists into early October. The model mirrors this prolonged event, capturing its intensity on Day-1, retaining the full class spectrum on Day-7, and reflecting the gradual decline in Extreme pixels by Day-14.}
\label{september_bloom}
\end{figure}

\paragraph{\textbf{St. Albans Bay}}
The time series visualization for St. Albans Bay (\Cref{st_hitrate_results_20232024}) reveals different CyanoHABs characteristics compared to Missisquoi Bay. The visualization highlights distinct temporal and ecological patterns in CyanoHAB dynamics compared to Missisquoi Bay. CyanoHABs development in St. Albans tends to occur later in the season, typically beginning around middle or late July, as indicated by the initial drop in hit rate during that period. This temporal shift likely reflects different environmental conditions, possibly related to varying warming patterns or nutrient dynamics between the two locations.

The CyanoHABs phase in St. Albans is characterized by fragmented and intermittent activity, reflected in the oscillating hit rate values throughout August and September. These fluctuations imply multiple short-lived or low-intensity CyanoHABs events, which challenge prediction continuity. Despite this, the model demonstrates consistent recovery in hit rate during sustained CyanoHABs phases, such as late September and early October for both years, where performance stabilizes with a higher hit rate.

Year-to-year differences are also notable. While 2023 exhibits more variable hit rates during the CyanoHABs season, 2024 shows a smoother and stable trajectory, indicating longer CyanoHABs events with more substantial alignment between model forecasts and observations. This pattern suggests that CyanoHABs drivers in 2024 may have been more consistent, enhancing forecast reliability at this site.

\paragraph{\textbf{Northeast Arm}}
The Northeast Arm, an oligotrophic zone characterized by relatively low nutrient levels and biological productivity, is divided into upper and lower sections for analysis due to their distinct CyanoHAB characteristics. The upper side is shown in \Cref{ne_upper_hitrate_results_20232024}, exhibits CyanoHABs that start forming in late August and persist intermittently through October. The consistent fluctuations in hit rate during these months for both 2023 and 2024 suggest frequent, short-duration, and fragmented CyanoHABs events, which present challenges for sustained predictive accuracy.

The lower side is depicted in \Cref{ne_lower_hitrate_results_20232024}, demonstrating a more stable CyanoHABs structure, particularly in late September 2023, as evidenced by the extended period of high hit rate. This suggests longer-lasting and more predictable CyanoHAB conditions during that period. However, 2024 presents a different pattern: the hit rate exhibits frequent fluctuations, indicating that CyanoHABs events in the lower region became more intermittent and fragmented, resembling the instability seen in the Upper Side. These variations highlight the interannual variability in CyanoHABs dynamics and environmental forcing.

The contrast between the two sub-regions, especially in 2023, demonstrates the benefit of spatial subdivision. While both sides are geographically adjacent, their temporal patterns, CyanoHABs persistence, and ecological responses differ meaningfully. This justifies the modeling decision to split the Northeast Arm, enabling more precise detection and characterization of localized CyanoHAB behavior.

\section{Discussion}
Remote sensing data are used to forecast the occurrence of CyanoHABs in Lake Champlain's three key segments: Missisquoi Bay, St. Albans Bay, and Northeast Arm, over a 14-day forecast horizon. The CI Values from the Cyanobacterial Assessment network indicate the intensity of CyanoHABs. Other data features include Temperature and temporal data features. The proposed research addresses two fundamental questions: 1) Given the nature of sparse remote sensing data, can we precisely forecast the CyanoHABs? 2) If we can forecast CyanoHABs early, can we also predict their intensity? 

The experimental results affirm both questions. The Transformer-BiLSTM can reliably predict the early occurrences of CyanoHABs, along with their intensities, across all five classes within a 14-day forecast horizon. Despite the inherent complexity of a 14-day forecasting window, the Transformer-BiLSTM consistently demonstrates seasonal CyanoHABs events and predictions alignments across different stations and years. The model captures both the onset and duration of CyanoHAB activity and maintains strong forecasting performance even under varying seasonal patterns and interannual conditions. Across all regions, the hit rate time series reflects the seasonal nature of CyanoHABs, with prediction accuracy adapting to site-specific CyanoHABs dynamics. For instance, the model detects early CyanoHAB formation in Missisquoi Bay during June 2024, captures intermittent events in St. Albans Bay, and differentiates between the fragmented CyanoHAB behavior of the Upper Northeast Arm and the more sustained patterns seen in the Lower Side. These results also validate the spatial subdivision of the Northeast Arm, where distinct hit rate profiles reveal different ecological responses within geographically proximate zones.

Analysis of prediction confidence scores provides additional insights into the system's performance, further validating the model's forecasting reliability. The model provides confidence scores that align with known CyanoHAB patterns, as shown in \Cref{mq_confidence_scores} for Missisquoi Bay. High confidence is observed during peak CyanoHAB periods across all intensity classes, whereas confidence levels are near zero during non-CyanoHAB seasons. These patterns demonstrate that the model has correctly learned the seasonal dynamics of CyanoHABs without overfitting. The apparent differences between 2023 and 2024 confidence patterns indicate that the model accurately captures year-to-year variations. During the CyanoHAB start and end periods, confidence scores differ between intensity classes, with low-intensity HABs exhibiting higher confidence than extreme events, reflecting the natural progression of CyanoHABs. These confidence patterns confirm that the model effectively forecasts CyanoHAB presence and intensity levels despite data sparsity.

Figures~\ref{mb_maps_1}–\ref{september_bloom} further validate this behavior by setting predictions following satellite observations for four representative CyanoHABs events. These diagrams illustrate the bin-classification framework used in this study (See \Cref{data_extraction}). Each row corresponds to a forecast lead time (Day-1, 7, 14). The five left-hand tiles show the predicted presence of CyanoHABs for the discrete CI-intensity bins, while the matching tiles beneath them display the satellite‐derived masks for the same day. The composite on the far right shows the complete Missisquoi Bay station CI file after performing the data imputation (See \Cref{data_sparsity}).  Since the problem is formulated at the bin level rather than at the true pixel scale, the model predicts each intensity bin and does not attempt sub-pixel localization. Figures~\ref{mb_maps_1}–\ref{mb_maps_2} show two July 2024 cases, the forecast assigns pixels to all five intensity classes at the Day-1 horizon and maintains a similar class composition through Day-14, aligning with the observed CyanoHABs events. The \Cref{model_underpredictions} illustrates a different behavior: the Day-1 forecast underestimates an abrupt flare-up, yet by Day-7 it reproduces the observed bloom magnitude and by Day-14 it represents the bay-wide distribution, indicating rapid model adjustment once larger-scale drivers prevail. The \Cref{september_bloom} shows a late-September to early-October sequence that depicts a high-intensity bloom that persists beyond the typical season; both forecast and satellite data retain High- and Extreme-class pixels into early October, implying that warmer surface waters and extended stratification are lengthening bloom lifetimes. 

\begin{figure*}
\centering
\includegraphics[width=1\textwidth, height=1\textwidth, keepaspectratio]{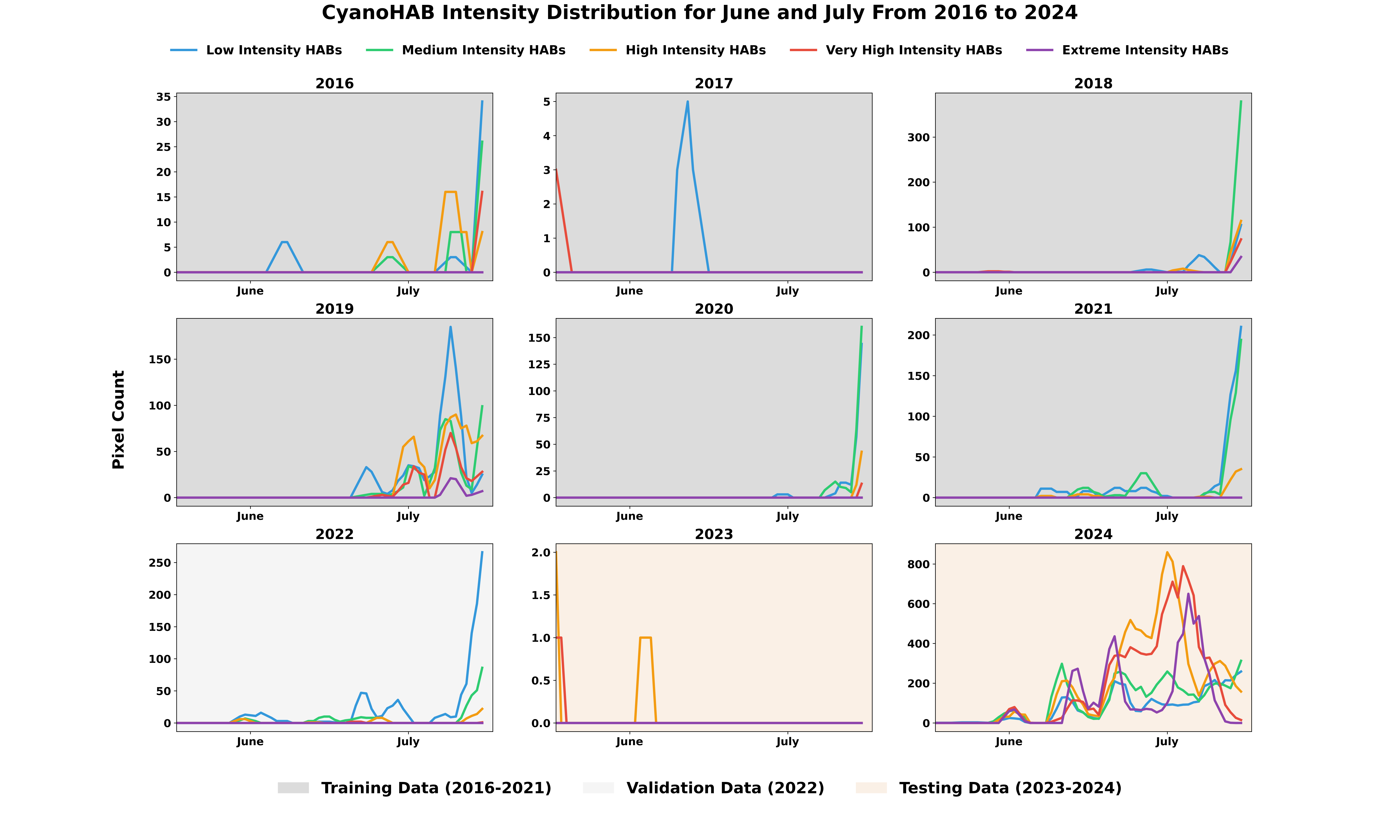}
\caption{CyanoHAB intensity distribution for June and July across all years from 2016 to 2024 in Missisquoi Bay, showing pixel counts for five intensity classes. Low is shown in blue, Medium in green, High in orange, Very High in red, and purple indicates Extreme Intensity CyanoHABs. Background colors indicate data partition: gray represents training data years, light gray represents validation years, and light orange represents testing years. One can observe the presence of CyanoHABs events in late July, with moderate or minimal pixel counts in June. The 2024 testing data shows a dramatic shift with unprecedented early-season blooms beginning in early June and exceptionally high pixel counts across all intensity classes.}
\label{mq_pixel_count}
\end{figure*}

Since six additional models were trained for performance evaluation against the Transformer-BiLSTM model (see \Cref{forecasting_quantitative_evaluation}), we evaluated an ensemble approach to assess potential performance gains. The ensemble method averaged predictions from all seven models to generate final forecasts. While the ensemble approach achieved marginally better performance with an F1 score of 79.09 compared to the Transformer-BiLSTM's F1 score of 78.86, the improvement was minimal at only 0.23 percentage points. However, ensemble implementation requires independent training and testing of each constituent model, significantly increasing computational complexity and deployment overhead. Given the marginal performance enhancement, the single Transformer-BiLSTM model presents a more practical solution for operational deployment, offering faster inference times and reduced computational requirements. 

The model also demonstrated generalization to unusual CyanoHAB conditions not present in the training data, particularly evident during the June and July 2024 period. As illustrated in \Cref{mq_pixel_count}, 2024 exhibited unprecedented early-season CyanoHAB events with exceptional intensity levels not observed during the training period. While training years showed CyanoHAB events predominantly in late July with moderate pixel counts, 2024 presented high-intensity events beginning in early June, with pixel counts higher than in any previous training year. This represents a significant departure from historical patterns, with hundreds of pixels across all intensity classes, including the extreme category. Despite never encountering such conditions during training, the model accurately forecast these anomalous events as confirmed in \Cref{mq_hitrate_results_20232024}. This capacity to predict CyanoHAB patterns, rather than merely reproducing historical observations, validates the model's ability to learn fundamental environmental relationships governing CyanoHAB formation, rather than simply memorizing seasonal templates.

These predictions of unusual events reveal spatial and temporal heterogeneity across Lake Champlain that presents substantial forecasting challenges. The unprecedented early-season CyanoHAB events observed in Missisquoi Bay in June 2024, with pixel counts exceeding historical patterns, suggest potential shifts in CyanoHAB dynamics that may be attributable to changing environmental conditions. Each monitoring segment shows distinct CyanoHAB patterns: persistent high-intensity events in Missisquoi Bay, episodic patterns in St. Albans Bay, and complex regional variations in Northeast Arm. These differences highlight the localized nature of CyanoHAB formation processes. Notably, the counterintuitive finding is that the lower side of the Northeast Arm, despite its oligotrophic designation and distance from the nutrient-rich zone of Missisquoi Bay, exhibited more stable events than the upper side in 2024. This suggests that complex hydrodynamic processes, beyond simple nutrient gradients, influence the development of CyanoHABs. The significant inter-annual variability observed between 2023 and 2024 across all segments further complicates forecasting efforts, with some locations showing earlier CyanoHAB onset, others displaying more substantial fragmentation, and all exhibiting distinct intensity distribution patterns. These findings demonstrate that effective CyanoHAB forecasting requires sophisticated algorithms and consideration of site-specific ecological dynamics.

\section{Conclusion}
This work utilizes remote sensing data to present a Transformer-BiLSTM framework for forecasting CyanoHAB occurrences and intensity levels at three key segments in Lake Champlain: Missisquoi Bay, St. Albans Bay, and Northeast Arm. The system is designed to provide day-by-day forecasts across a 14-day horizon, offering fine-grained temporal insights into the dynamics of CyanoHABs. The research uses two remote sensing data features: Cyanobacterial Index values and Temperature. The system addresses data sparsity through a two-stage preprocessing pipeline combining forward fill and weighted temporal imputation, enabling effective forecasting despite significant missing data in both CI values and temperature measurements. A temporal data augmentation strategy is also presented to address the limited training dataset while preserving physically plausible relationships.

The model demonstrated generalization to novel CyanoHAB patterns, including unprecedented early-season high-intensity events in 2024 that were not represented in the training data. Analysis across different monitoring segments reveals distinct spatial characteristics: persistent high-intensity CyanoHABs in Missisquoi Bay, episodic patterns in St. Albans Bay, and complex regional variations in Northeast Arm. These segment-specific patterns and significant inter-annual variability between 2023 and 2024 highlight the importance of localized approaches for effective CyanoHAB management. Compared to a persistence baseline, which performed better on day 1, the Transformer-BiLSTM consistently outperformed across the 14-day horizon, with a 7.62 percentage point advantage by day 14, confirming its robustness in capturing the evolving behavior of CyanoHABs. Comparative evaluation against seven advanced deep learning architectures further validated the Transformer-BiLSTM's performance for extended forecasting horizons. While Transformer-only models demonstrated competitive performance in short-term predictions, the Transformer-BiLSTM architecture outperformed them for longer-duration forecasting tasks. Prediction confidence scores provided further validation of the model's interpretability. High confidence aligned with peak CyanoHAB periods across all intensity classes, while confidence diminished in off-season intervals. Inter-annual variations in confidence patterns between 2023 and 2024 reflected the model’s sensitivity to year-specific dynamics without overfitting. 

The proposed framework effectively forecasts CyanoHAB occurrences and intensity levels over a 14-day horizon. However, we acknowledge potential areas for future enhancement. Current predictions provide segment-level resolution; moving toward pixel-level localization will improve spatial specificity. The reliance on MODIS temperature data poses a challenge due to its planned decommissioning, which necessitates transitioning to alternative sources like Visible Infrared Imaging Radiometer Suite, or Sentinel-3 Sea and Land Surface Temperature Radiometer. Future work addresses these limitations by developing a pixel-level forecasting system and integrating additional satellite-based input features to ensure continuity. 

\section*{Funding}
Funding for this project was provided by the National Oceanic and Atmospheric Administration, which was awarded to the Cooperative Institute for Research on Hydrology through the NOAA Cooperative Agreement with The University of Alabama, NA22NWS4320003.

\section*{Conflict of Interest Statement}
The authors declare that they have no known competing financial interests or personal relationships that could have influenced the work reported in this paper.

\bibliographystyle{IEEEtran}
\bibliography{References}

\end{document}